\documentclass[11pt]{article}

\usepackage[final]{latex/acl}

\usepackage{times}
\usepackage{latexsym}
\usepackage[T1]{fontenc}
\usepackage[utf8]{inputenc}
\usepackage{microtype}
\usepackage{inconsolata}
\usepackage{graphicx}

\usepackage{amsmath}
\usepackage{amssymb}
\usepackage{amsfonts}
\usepackage{booktabs}
\usepackage{multirow}
\usepackage{adjustbox}
\usepackage{arydshln}
\usepackage{colortbl}
\usepackage{xcolor}
\usepackage{pifont}
\usepackage{makecell}
\usepackage{caption}
\usepackage{subcaption}
\usepackage{wrapfig}
\usepackage{enumitem}
\usepackage{xspace}
\usepackage{tabulary}
\usepackage{threeparttable}
\usepackage[most]{tcolorbox}
\usepackage{algorithm}
\usepackage{algpseudocode}
\usepackage{placeins}

\definecolor{Gray}{gray}{0.5}
\definecolor{LGray}{gray}{0.9}
\definecolor{darkblue}{RGB}{0, 0, 139}
\definecolor{lightblue}{RGB}{173, 216, 230}
\definecolor{myblue}{HTML}{86B0C6}
\definecolor{blue2}{HTML}{7497BE}
\definecolor{myyellow}{HTML}{E7805E}
\definecolor{yellow2}{HTML}{FBC98C}
\definecolor{mygreen}{HTML}{6DAF90}
\definecolor{highlightgreen}{HTML}{C7E6D3}
\definecolor{lightred}{HTML}{D32F2F}
\definecolor{reasoncolor}{RGB}{30, 90, 170}
\definecolor{latentcolor}{RGB}{200, 110, 30}
\definecolor{answercolor}{RGB}{40, 130, 70}
\definecolor{boxframe}{HTML}{64748B}
\definecolor{boxback}{HTML}{F1F5F9}
\colorlet{Budapest}{teal!15}
\colorlet{budapest}{teal!15}

\def\MODEL{Future-L1}
\newcommand{\method}{\textsc{\MODEL}\xspace}

\newcommand{\modelSFT}{\textsc{\MODEL-SFT}\xspace}
\newcommand{\modelRL}{\textsc{\MODEL-RL}\xspace}
\newcommand{\dataset}{\textsc{\MODEL-50K}\xspace}

\makeatletter
\newcommand\blfootnote[1]{%
  \begingroup
  \renewcommand\thefootnote{}\footnote{#1}%
  \addtocounter{footnote}{-1}%
  \endgroup
}
\makeatother

\title{Imagine Before You Predict: Interleaved Latent Visual Reasoning \\for Video Event Prediction}

\author{
Tianxiang Jiang$^{\ast1,2}$ \quad 
Linquan Wu$^{\ast3}$ \quad
Sheng Xia$^{4}$ \quad
Songze Li$^{5,2}$ \\
\textbf{Ziang Yan$^{6,2}$} \quad 
\textbf{Haoyu Yang$^{7}$} \quad
\textbf{Yu Qiao$^{2}$} \quad 
\textbf{Yi Wang$^{2\dagger}$} \\ \\
$^{1}$ University of Science and Technology of China \quad
$^{2}$ Shanghai AI Laboratory \\
$^{3}$ City University of Hong Kong \quad
$^{4}$ Nanjing University \quad
$^{5}$ Fudan University
\\
$^{6}$ Zhejiang University \quad
$^{7}$ University of Electronic Science and Technology of China \\
\href{https://github.com/OpenGVLab/Future-L1}{\textcolor{reasoncolor}{\texttt{\textbf{https://github.com/OpenGVLab/Future-L1}}}}
}

\begin{document}

\maketitle

% \renewcommand*{\thefootnote}{}
% \footnotetext[1]{$^\dagger$Corresponding Author when at Shanghai AI Lab. \\ $^\ast$Equal Contribution.}

\begin{abstract}
Video event prediction (VEP) requires models to infer unobserved future states from partial video evidence. Existing video MLLMs usually verbalize intermediate future reasoning in text space: once visual evidence is verbalized, fine-grained motion, geometry, and interaction cues can be lost, leading to plausible but visually ungrounded hallucinations. We introduce \textbf{\method}, an interleaved latent visual reasoning framework that lets an MLLM alternate between language tokens and continuous latent visual spans during autoregressive decoding. To train this capability, we construct \textbf{\dataset} by selecting examples where future visual hints help prediction and align latent states to future-frame embeddings, then further optimize sampled latent trajectories with \textbf{LA-DAPO}, a latent-aware RL objective with outcome-contrastive and temporal-diversity rewards. \method achieves new \textit{state-of-the-art} results on both benchmarks: on FutureBench, it improves Qwen3-VL-8B from 61.0 to \textbf{85.4} and exceeds the previous best Video-CoE by 10.4 points; on TwiFF-Bench, it improves the average score from 2.44 to \textbf{3.04}. These results suggest that future-oriented video reasoning benefits from preserving intermediate visual semantics in latent space rather than translating every reasoning step into text.

\blfootnote{$\dagger$ Corresponding author. This work was performed at Shanghai AI Laboratory. $*$ Equal contribution.}
% \blfootnote{}

\end{abstract}

\section{Introduction}
\label{sec:Introduction}

\begin{figure*}[t]
    \centering
    \includegraphics[width=\linewidth]{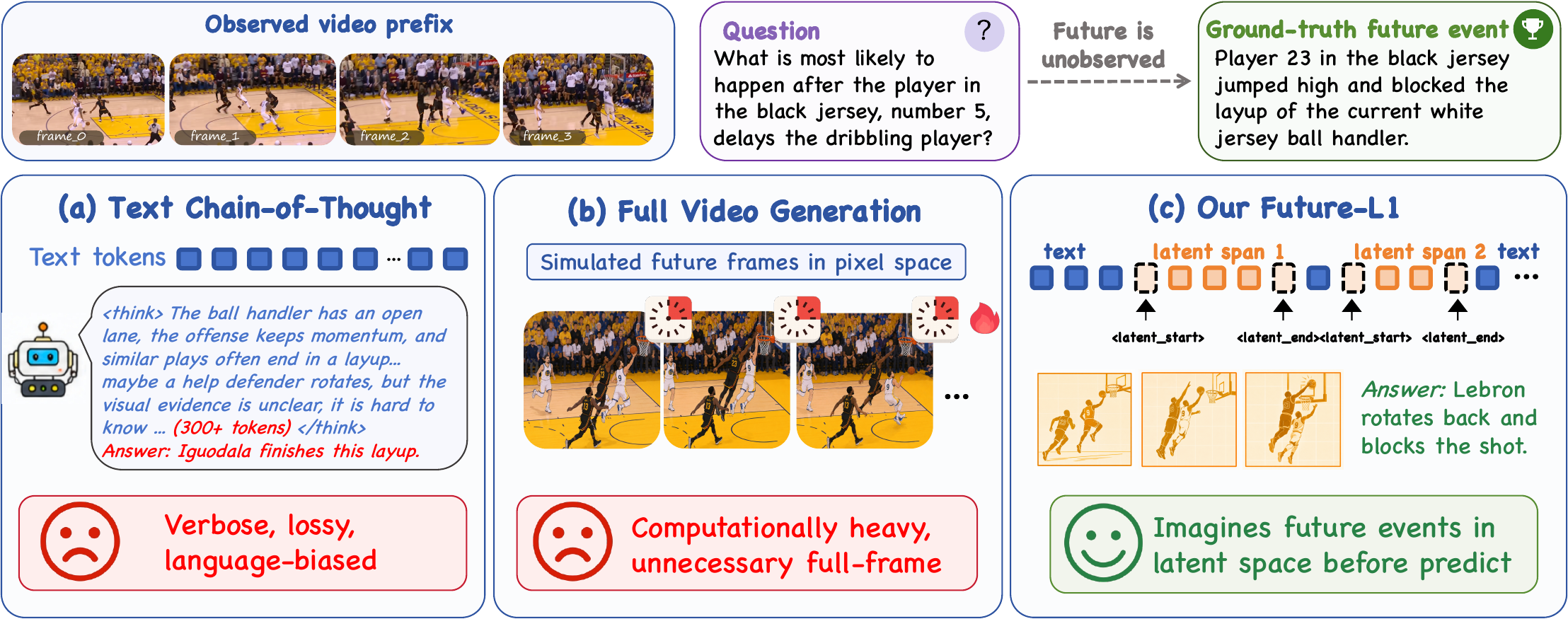}
    % \vspace{-5mm}
    \caption{
    \textbf{Motivation of interleaved latent visual reasoning.} Text-CoT can be verbose and visually lossy, while pixel-space future simulation is computationally heavy. \textbf{\method} instead inserts compact latent visual spans that preserve dynamic future semantics without generating full frames.}
    \vspace{-2mm}
    \label{fig:introduction}
\end{figure*}

Video event prediction (VEP) asks a model to infer what will happen next from a partially observed video~\cite{7102751,vondrick2016anticipating,lei2020more,wang2025fostering,su2026video}. Unlike standard video understanding, whose answers can usually be grounded in visible frames, VEP requires constructing an internal hypothesis about unobserved dynamic visual states: where objects will move, whether entities will interact, and how a scene will evolve. Although recent multimodal large language models (MLLMs) have made rapid progress on retrospective video tasks~\cite{bai2025qwen25vltechnicalreport,bai2025qwen3,wang2024internvideo2,li2024mvbench,fu2025videommefirstevercomprehensiveevaluation,li2025learning}, future-oriented reasoning remains less explored.

Existing video MLLMs usually verbalize intermediate future reasoning in text space~\cite{zhang2023multimodal,han2025videoespresso,feng2025videor1reinforcingvideoreasoning,li2025videochatr1,su2026video}. This is convenient for explanation, but it creates a poor interface for dynamic visual prediction: once visual evidence is converted into words, fine-grained motion, geometry, relative position, and interaction can be lost. The resulting reasoning may sound plausible while drifting away from visual semantics, especially when the correct answer depends on subtle future dynamics. Recent latent visual reasoning methods avoid part of this bottleneck by using continuous visual states~\cite{li2025mvot,pham2025mcout,qin2025covt,10.1007/978-3-032-37592-6_35,li2025latent,yang2025machine,lu2026onevl}, but most treat latent thoughts as static helper images or one-shot visual hints. VEP instead calls for a temporally organized latent process that can update imagined dynamic visual states over multiple reasoning steps.

We introduce \textbf{\method}, a framework that equips MLLMs with \textbf{interleaved latent visual reasoning} for VEP. During autoregressive decoding, \method{} alternates between textual tokens and continuous latent visual spans, allowing language to organize the reasoning while latent states preserve intermediate dynamic visual structure. Training proceeds in two stages. First, we construct \textbf{\dataset} from TwiFF-style trajectories using visual-gain data curation, selecting examples where intermediate future visual hints measurably help prediction. Supervised fine-tuning then teaches the model when to invoke latent spans and aligns their hidden states with future-frame embeddings. Second, we apply \textbf{LA-DAPO}, a latent-aware RL objective that optimizes sampled latent trajectories with outcome-contrastive and temporal-diversity rewards, encouraging successful latent futures while discouraging repeated visual thoughts.

Experiments show that latent visual reasoning is substantially more effective than text-only reasoning for VEP. On FutureBench, \modelRL{} improves Qwen3-VL-8B from 61.0 to \textbf{85.4}, exceeding the previous best Video-CoE by \textbf{10.4} points. On TwiFF-Bench, it improves the average score from 2.44 to \textbf{3.04}. Under the same curated data source, text-only SFT reaches only 65.0 on FutureBench, whereas interleaved latent SFT reaches \textbf{73.2}, indicating that the gain is not merely from additional supervision but from reasoning through a modality better matched to future visual structure.

Our contributions are threefold:
\begin{enumerate}[leftmargin=*,itemsep=2pt,topsep=2pt]
\item We propose visual-gain data curation and construct \textbf{\dataset}, a high-utility corpus for supervising latent future visual reasoning.
% \vspace{-2.5mm}
\item We introduce \textbf{interleaved latent visual reasoning for VEP}, enabling autoregressive models to alternate between language and continuous future visual states.
% \vspace{-2.5mm}
\item We develop \textbf{LA-DAPO}, a latent-aware RL method that improves sampled latent trajectories and achieves \textbf{state-of-the-art} results on FutureBench and TwiFF-Bench.
\end{enumerate}

\section{Related Work}
\label{sec:Related_work}

\paragraph{Multimodal Large Language Models.}
Multimodal large language models (MLLMs) connect visual encoders with strong LLM backbones and have become the mainstream framework for visual understanding~\cite{bai2025qwen3,team2026kimi,hong2026glm,xiao2026mimo,an2026llavaonevision2nextgenerationperceptualintelligence}.
For video understanding, recent MLLMs extend image-based models with temporal frame sampling, video instruction tuning, longer-context modeling, and large-scale video-text corpora~\cite{wang2024internvideo2,zhang2024llava,wang2025make, yan2026internvideo3}, substantially improving performance on diverse benchmarks~\cite{li2024mvbench,fu2026video,yang2025thinking,xu2025expvid,shi2026river,liu2026ovibenchbenchmarkingonlinevideo}.
Beyond perception and recognition, reasoning-oriented post-training has been applied to MLLMs, including chain-of-thought supervision~\cite{han2025videoespresso} and reinforcement learning~\cite{li2025videochatr1}.
More recently, paradigms that encourage models to \textit{think with images or videos} move beyond purely textual rationales by retrieving visual evidence~\cite{zheng2025deepeyes,zeng2026video,wang2026see, lu2026think, wu2026see} with visual traces, motivating non-textual intermediate representations for visual reasoning.

\paragraph{Reasoning in Latent Space.}
Latent reasoning~\cite{yu2026latent} replaces discrete textual reasoning tokens with continuous hidden states fed back into the LLM, compressing chain-of-thought into a compact thinking space.
Coconut~\cite{hao2024training} first showed that an LLM can reason in its own embedding space, and CODI~\cite{shen2025codi} and SIM-CoT~\cite{wei2025sim} subsequently distilled or supervised these latent steps to close the gap to explicit textual CoT.
This paradigm has also been adopted by MLLMs through visual supervision: Mirage~\cite{yang2025machine} and LVR~\cite{li2025latent} align latent slots with embeddings of helper images that hint at the answer, and LaViT~\cite{wu2026lavit} further constrains latent visual thoughts with teacher-guided attention.
More flexible designs allow models to alternate between textual tokens and continuous visual states during reasoning, as in Monet~\cite{wang2025monet}, SkiLa~\cite{tong2025sketch}, and SwimBird~\cite{tong2026swimbird}.
However, these methods largely anchor latent thoughts to \emph{static images}, such as helper images, sketches, or scenes already given to the model. Video event prediction instead requires reasoning over \emph{dynamic future frames} that are not yet observed, where above studies have not explored. \method{} accordingly grounds latent thoughts in future information rather than static visual hints.

\paragraph{Video Event Prediction.}
Unlike standard video understanding benchmarks~\cite{li2024mvbench,fu2025videommefirstevercomprehensiveevaluation,liu2024tempcompass} that focus on visible content, video event prediction requires models to infer unobserved future events from a video prefix. This future-oriented setting spans low-level action anticipation~\cite{10.1007/978-3-319-10578-9_45,Gammulle_2019_ICCV}, future-frame prediction~\cite{ranzato2014video,vondrick2016generating}, and high-level semantic next-event prediction~\cite{lei2020more,jiang2025vknowu,liang2025videventlargedatasetunderstanding,su2025eventformer}.
Most VEP methods remain text-output oriented~\cite{cheng2025tempura,wang2025fostering}; for example, Video-CoE~\cite{su2026video} structures the reasoning trace as a long textual chain of historical events. Video-as-Answer~\cite{cheng2025video} instead moves the answer modality from text to generated video explicitly.
\method{} differs from these routes: rather than verbalizing every intermediate event or synthesizing full videos, it represents intermediate future states in an interleaved latent visual channel supervised by future-frame embeddings.

\begin{figure*}[t]
    \centering
    \includegraphics[width=\textwidth]{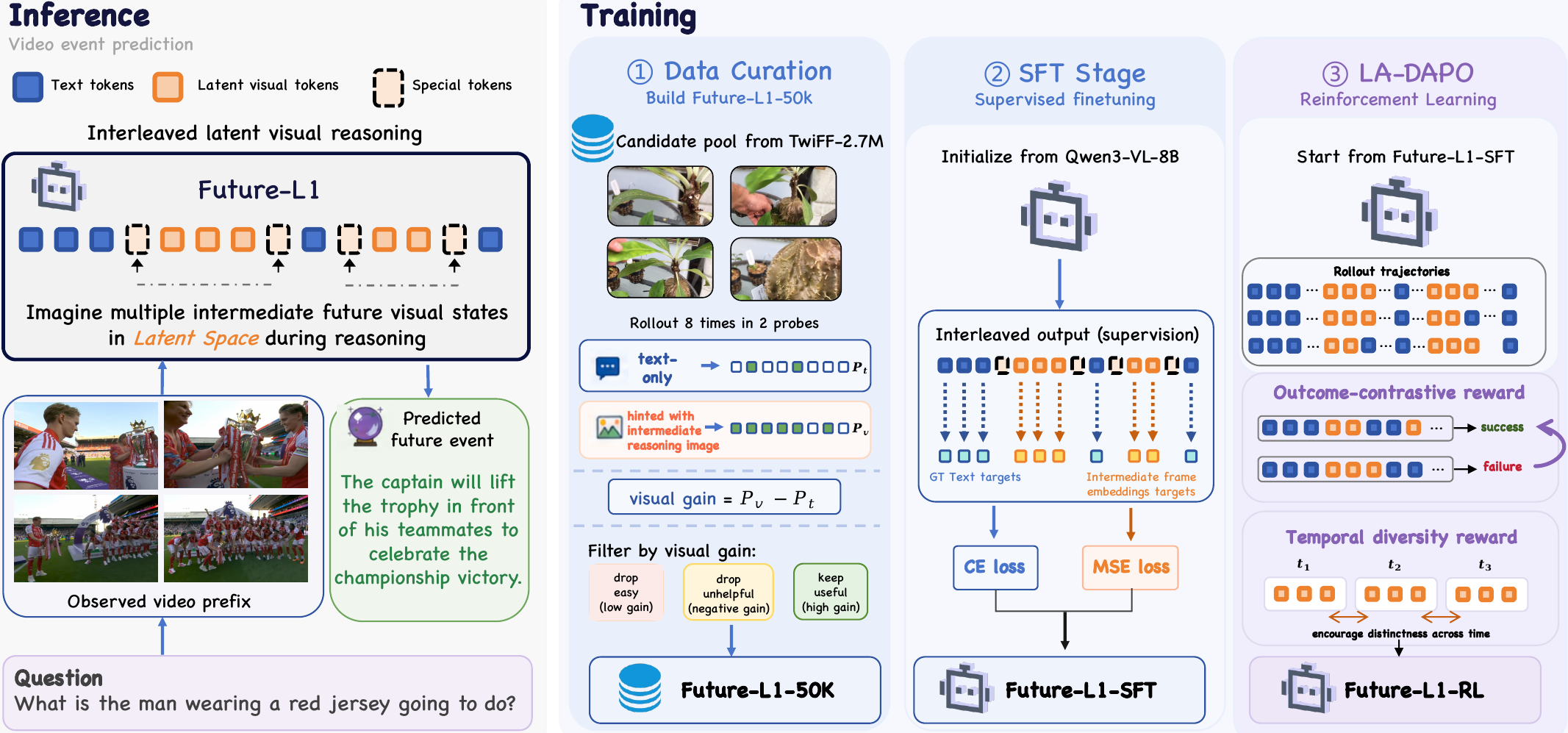}
    % \vspace{-5mm}
    \caption{
    \textbf{Overview of \method.} (Left) \dataset{} is built by ranking TwiFF candidates by visual gain $p_v-p_t$. (Center) SFT trains interleaved text--latent trajectories, aligning latent spans with future visual states. (Right) LA-DAPO further optimizes sampled trajectories with outcome-contrastive and temporal-diversity rewards.}
    \vspace{-2mm}
    \label{fig:architecture}
\end{figure*}

\section{Method}
\label{sec:method}

% \subsection{Overview}
% \label{ssec:overview}

We propose \method{}, an interleaved latent visual reasoning framework for VEP. Given an observed video prefix $V$ and question $q$, the model generates a response $y$ by alternating textual reasoning, bounded latent visual spans, and a final answer. Training has two stages: SFT on \dataset{} teaches when to invoke latent spans and aligns them with future-frame embeddings, while LA-DAPO further optimizes sampled latent trajectories with outcome-contrastive and temporal-diversity rewards. Figure~\ref{fig:architecture} illustrates the pipeline.

\subsection{Interleaved Latent Visual Reasoning}
\label{ssec:prelim}

\paragraph{Autoregressive Reasoning with Latent Visual Spans.}
\method{} augments a standard MLLM backbone~\cite{bai2025qwen3} with a latent visual reasoning channel using three special tokens: \textcolor{latentcolor}{\textbf{\texttt{<|latent\_start|>}}}, \textcolor{latentcolor}{\textbf{\texttt{<|latent|>}}}, and \textcolor{latentcolor}{\textbf{\texttt{<|latent\_end|>}}}. Generation begins in textual mode. Once \texttt{<|latent\_start|>} is emitted, each following \texttt{<|latent|>} position produces a hidden state $\mathbf{h}_t$ that is fed back as the next input embedding rather than projected to the vocabulary. These continuous states act as latent visual thoughts and remain in the KV cache to condition later textual reasoning. Generation returns to text when \texttt{<|latent\_end|>} is emitted.

\paragraph{Dynamic Latent Budget at Inference.}
Latent span length is not fixed: a span ends when the model emits \texttt{<|latent\_end|>}. We cap each span by $L_{\max}$ to avoid run-on latent decoding, and a response may contain multiple spans, allowing the model to allocate latent computation adaptively across reasoning stages.

\subsection{SFT with \dataset}
\label{ssec:sft50k}
SFT provides a necessary cold start for latent reasoning by training on curated interleaved traces and aligning latent states with future-frame embeddings. This prevents the model from either avoiding latent spans or producing continuous states not grounded in meaningful visual manifold before RL.

\paragraph{Visual-Gain Data Curation.}
We curate \textbf{\dataset} from TwiFF-2.7M~\cite{liu2026twiff}, a VCoT corpus that provides intermediate reasoning frames. Unlike synthesized sketches or generic helper images, these frames are temporally later frames from the same authentic video, so they depict unseen future states that are physically consistent with the observed prefix. This makes them a natural supervision signal for latent visual reasoning: the model is not asked to imitate arbitrary visual hints, but to internalize future visual states that actually occur. 

However, not every TwiFF sample provides useful supervision for VEP. Some examples are already easy to solve from the observed prefix alone, where extra future-frame hints add little value. Others remain ambiguous or uninformative even when a reasoning frame is provided. Training on them dilutes the signal that latent visual states should carry.
We therefore filter examples by the \emph{marginal utility} of their intermediate reasoning frames.

For each candidate, we evaluate Qwen3-VL-8B-Instruct under two conditions: (1) a text-only input with the observed video prefix and question; and (2) a hinted input that additionally includes the intermediate reasoning frames. Each condition uses 8 independent rollouts judged by Qwen3.5-397B-A17B. Let $p_t, p_v \in [0,8]$ be the correct-rollout counts; we \textbf{retain} samples with $p_t \le 6$, so the text-only setting is not saturated, and $p_v - p_t \ge 2$, so the visual hint provides measurable lift. We \textbf{rank} retained samples by descending $p_v - p_t$, and take the top 50{,}000 items as \textbf{\dataset}. All retained samples are reformatted into the interleaved trajectory shown in Figure~\ref{fig:training_sample}.

\begin{figure}[t]
\centering
\begin{tcolorbox}[
  colback=white,
  colframe=boxframe,
  colbacktitle=boxback,
  coltitle=black,
  arc=2pt,
  boxsep=2pt,
  left=6pt,right=6pt,top=4pt,bottom=4pt,
  title={\small \textbf{\dataset{} Training Example}},
  fonttitle=\bfseries
]
\scriptsize\ttfamily
% [Visual Prefix $V$] \quad [Question $Q$] \\[2pt]
\textcolor{reasoncolor}{\texttt{<reason>} [Textual CoT 0] \texttt{</reason>}} \\
\textcolor{latentcolor}{\texttt{<|latent\_start|>} [Latent Visual Frames 1] \texttt{<|latent\_end|>}} \\
\textcolor{reasoncolor}{\texttt{<reason>} [Textual CoT 1] \texttt{</reason>}} \\
\dots \dots \\
\textcolor{latentcolor}{\texttt{<|latent\_start|>} [Latent Visual Frames N] \texttt{<|latent\_end|>}} \\
\textcolor{reasoncolor}{\texttt{<reason>} [Textual CoT N] \texttt{</reason>}} \\
\textcolor{answercolor}{\texttt{<answer>} [Predicted Future Event] </answer>}
\end{tcolorbox}
\vspace{-0.3em}
\caption{\dataset{} training format: textual reasoning interleaved with bounded latent visual spans supervised by future-frame embeddings.}
\vspace{-3mm}
\label{fig:training_sample}
\end{figure}

\vspace{-2mm}
\paragraph{Training Objective.}
SFT optimizes a joint objective over discrete text tokens and continuous latent visual states:
\begin{equation}
\mathcal{L}_{\text{SFT}} = \mathcal{L}_{\text{CE}} + \lambda \mathcal{L}_{\text{Latent}},
\label{eq:loss}
\end{equation}
where \(\lambda\) controls the strength of latent supervision.
% and is ablated in \S\ref{sec:Experiments}.

For discrete positions \(\mathcal{T}\), including textual reasoning, answer tokens, and special control tokens, we use standard next-token prediction:
\begin{equation}
\mathcal{L}_{\text{CE}} = -\sum_{t \in \mathcal{T}} \log p_\theta\!\left(w_t \mid w_{<t}, V, q \right).
\label{eq:ce}
\end{equation}

For latent positions \(\mathcal{S}\), we align each hidden state \(\mathbf{h}_t\) with the visual embedding \(\mathbf{e}_t^\star\) of the corresponding future reasoning frame, extracted by the Qwen3-VL vision encoder:
\begin{equation}
\mathcal{L}_{\text{Latent}} =
\frac{1}{|\mathcal{S}|}\sum_{t \in \mathcal{S}}
\big\| \mathbf{h}_t - \mathbf{e}_t^{\star} \big\|_2^2 .
\label{eq:lat}
\end{equation}
This anchors latent spans to the future-frame manifold while preserving standard language modeling over the textual channel.

\subsection{LA-DAPO for Latent-Aware RL}
\label{ssec:rl}

SFT provides a grounded but teacher-forced initialization: each latent state is matched to a future-frame embedding, while sampled latent trajectories are not directly optimized for prediction success. We therefore introduce \textbf{LA-DAPO} (\textbf{L}atent-\textbf{A}ware \textbf{D}irect \textbf{A}dvantage \textbf{P}olicy \textbf{O}ptimization), a latent-aware extension of DAPO~\cite{yu2025dapoopensourcellmreinforcement}. LA-DAPO keeps DAPO's answer and format rewards, and adds two trajectory-level latent rewards: an \textbf{outcome-contrastive reward} that aligns latent trajectories associated with correct answers, and a \textbf{temporal-diversity reward} that discourages repeating the same visual thought across spans. Because these rewards depend on rollout outcomes and generated latent states, LA-DAPO can optimize latent reasoning \textbf{without requiring intermediate-frame annotations} during RL.

\paragraph{Outcome-Contrastive Latent Reward.}
\label{ssec:contrastive_reward}
Answer rewards provide only a sequence-level scalar, leaving latent states weakly supervised. We introduce an \textbf{outcome-contrastive reward} \(R_{\mathrm{ctr}}\) that structures latent trajectories by group outcomes: correct rollouts are pulled together, while incorrect rollouts serve as negatives. Because the signal depends only on final-answer correctness, it does not require intermediate-frame annotations.

Let \(\mathbf{Z}_i=[\mathbf{z}_{i,1},\ldots,\mathbf{z}_{i,T_i}]\) be the normalized latent trajectory of rollout \(i\), with correctness \(a_i\in\{0,1\}\). We define trajectory similarity as
\begin{equation}
s_{ij}=\frac{1}{T}\sum_{t=1}^{T}\frac{1+\langle\mathbf{z}_{i,t},\mathbf{z}_{j,t}\rangle}{2},
\end{equation}
where \(T=\min(T_i,T_j)\). Let \(\mathcal{P}_i=\{j\ne i:a_j=1\}\), \(\mathcal{N}_i=\{j\ne i:a_j=0\}\), and \(s_i^+=\max_{j\in\mathcal{P}_i}s_{ij}\). We use a hardest-positive InfoNCE reward:
\begin{equation}
R_{\mathrm{ctr}}(i)
=
\frac{\exp(s_i^{+}/\tau)}
{\exp(s_i^{+}/\tau) + \sum_{j\in\mathcal{N}_i}\exp(s_{ij}/\tau)} .
\label{eq:ctr}
\end{equation}
Empty-positive or empty-negative cases are handled in Appendix~\ref{app:ladapo}.

\paragraph{Temporal Diversity Reward.}
\label{ssec:diversity_reward}
$R_{\mathrm{ctr}}$ aligns trajectories \emph{across} rollouts but imposes no structure \emph{within} a rollout: a policy can still earn a high answer reward by emitting near-identical latent states at consecutive spans, collapsing the latent channel into a single visual thought repeated over time. Although SFT discourages this through frame-distinct supervision, this constraint is no longer present during RL. We therefore add a \textbf{temporal diversity reward} \(R_{\mathrm{div}}\) that encourages adjacent latent spans to represent distinct future updates. 
For a response with \(M\) latent spans, we mean-pool the latent vectors within span \(m\) into a representative \(\mathbf{b}_m\), and penalize adjacent-span similarity:
\begin{equation}
R_{\mathrm{div}}
=
- \frac{1}{M-1}\sum_{m=1}^{M-1}\cos^2(\mathbf{b}_m,\mathbf{b}_{m+1}).
\label{eq:div}
\end{equation}
This reward is maximized at 0 when adjacent span representatives are orthogonal and decreases as they become redundant. 

Together, \(R_{\mathrm{ctr}}\) and \(R_{\mathrm{div}}\) regularize latent reasoning along complementary axes: \(R_{\mathrm{ctr}}\) links latent trajectories to prediction outcomes across rollouts, while \(R_{\mathrm{div}}\) keeps successive latent spans within a rollout temporally distinct.

\paragraph{Final Rewards.}
The total target combines answer / format rewards and two latent terms,
\begin{equation}
R
=
\lambda_a R_{\mathrm{acc}}
+ \lambda_f R_{\mathrm{fmt}}
+ \lambda_c R_{\mathrm{ctr}}
+ \lambda_d R_{\mathrm{div}} ,
\label{eq:rl_reward}
\end{equation}
where $\lambda_a$ and $\lambda_f$ follow the DAPO defaults (Appendix~\ref{app:ladapo}); $\lambda_c$ and $\lambda_d$ are ablated in \S\ref{sec:Experiments}.

% Starting from \modelSFT{}, LA-DAPO produces the final model, \textbf{\modelRL}. Intuitively, Stage 1 teaches the model to generate visually meaningful latent thoughts, while Stage 2 teaches it which latent trajectories are actually useful for correct future prediction.
\section{Experiments}
\label{sec:Experiments}

\begin{table*}[t]
\centering
\small
\setlength{\tabcolsep}{3.5pt}
\renewcommand{\arraystretch}{1.12}
\caption{\textbf{Main results on FutureBench}~\cite{wang2025fostering}. Accuracy (\%); best results are in \textbf{bold}.}
\label{tab:futurebench}
\resizebox{.95\textwidth}{!}{
\begin{tabular}{l|c|c|c|ccccc}
\toprule
\textbf{Model} & \textbf{Size} & \textbf{Method} & \textbf{Frames} & \textbf{1-Hop} & \textbf{2-Hop} & \textbf{3-Hop} & \textbf{Interp.} & \textbf{AVG} \\
\midrule
\multicolumn{9}{l}{\textit{Open-source and Proprietary Models}} \\
GLM-4.1V~\cite{vteam2026glm45vglm41vthinkingversatilemultimodal} & 9B & \multirow{10}{*}{Zero-Shot} & 32 & 29.9 & 41.9 & 52.2 & 47.3 & 44.4 \\
LLaVA-NeXT-Video~\cite{zhang2025llavavideovideoinstructiontuning} & 7B &  & 32 & 48.8 & 49.3 & 40.0 & 44.4 & 45.2 \\
% Kimi-VL-A3B & A3B &  & 32 & 44.3 & 42.8 & 51.3 & 51.9 & 48.9 \\
MiMo-VL~\cite{coreteam2025mimovltechnicalreport} & 7B &  & 32 & 59.0 & 59.6 & 50.5 & 43.8 & 50.5 \\
InternVL3~\cite{zhu2025internvl3} & 8B &  & 32 & 54.3 & 58.0 & 63.2 & 54.4 & 56.7 \\
% Qwen2.5-VL-Instruct & 3B &  & 32 & 66.5 & 62.7 & 63.2 & 55.2 & 46.3 \\
Qwen2.5-VL-Instruct~\cite{bai2025qwen25vltechnicalreport} & 7B &  & 32 & 57.2 & 57.0 & 50.2 & 50.7 & 52.9 \\
% Qwen2.5-VL-Instruct & 32B &  & 32 & 66.5 & 62.7 & 63.2 & 55.2 & 59.9 \\
Qwen2.5-VL-Instruct~\cite{bai2025qwen25vltechnicalreport} & 72B &  & 32 & 55.5 & 68.4 & 63.7 & 53.2 & 58.3 \\
Qwen3-VL~\cite{bai2025qwen3} & 30B-A3B &  & 32 & 65.3 & 70.5 & 76.1 & 62.2 & 66.9 \\
GPT-4o~\cite{openai2024gpt4o} & -- &  & 32 & 61.9 & 61.7 & 72.1 & 51.6 & 59.0 \\
GPT-5~\cite{singh2026openaigpt5card} & -- &  & 32 & 59.6 & 57.3 & 62.6 & 55.6 & 57.9 \\
\midrule
\multicolumn{9}{l}{\textit{Video Reasoning Models}} \\
Video-RFT~\cite{wang2025videorftincentivizingvideoreasoning} & 7B & SFT+RL & 32 & 62.4 & 53.9 & 50.7 & 53.8 & 54.6 \\
Video-R1~\cite{feng2025videor1reinforcingvideoreasoning} & 7B & SFT+RL & 32 & 67.6 & 65.3 & 61.2 & 61.8 & 63.3 \\
% VideoKnow+~\cite{jiang2025vknowu} & 7B & SFT+RL & 32 & -- & -- & -- & -- & -- \\
VideoAuto-R1~\cite{liu2026videoauto} & 8B & SFT+RL & 32 & 63.6 & 69.4 & 67.7 & 59.3 & 63.4 \\
Video-o3~\cite{zeng2026video} & 7B & SFT+RL & 32 & 68.2 & 73.6 & 63.2 & 69.7 & 68.9 \\
% NEP~\cite{wang2025fostering} & 7B & SFT & 32 & 57.2 & 57.0 & 50.2 & 50.7 & 52.9 \\
NEP~\cite{wang2025fostering} & 7B & SFT+RL & 32 & 66.2 & 69.9 & 63.7 & 68.1 & 67.3 \\
% Video-CoE~\cite{su2026video} & 7B & SFT & 32 & 67.6 & 74.1 & 62.2 & 63.2 & 65.7 \\
Video-CoE~\cite{su2026video} & 7B & SFT+RL & 32 & 80.9 & 83.9 & 71.6 & 71.4 & 75.0 \\
% VideoChat-R1~\cite{li2025videochatr1} & 7B & RL & -- & -- & -- & -- & -- & 46.6 \\
% Video-As-Answer~\cite{cheng2025videoasanswerpredictgeneratevideo} & -- & RL & -- & -- & -- & -- & -- & -- \\
\midrule
\multicolumn{9}{l}{\textit{Latent Visual Reasoning Models}} \\
LVR~\cite{li2025latent} & 7B & SFT+RL & 32 & 22.5 & 26.4 & 22.9 & 17.6 & 21.0$^{\dagger}$ \\
Monet~\cite{wang2025monet} & 7B & SFT+RL & 32 & 46.8 & 47.2 & 45.3 & 49.7 & 47.9 \\
SwimBird~\cite{tong2026swimbird} & 8B & SFT & 32 & 59.0 & 66.8 & 64.7 & 61.8 & 62.8 \\
\midrule
\multicolumn{9}{l}{\textit{Ours}} \\
Qwen3-VL-Instruct~\cite{bai2025qwen3} & 8B & Zero-Shot & 32 & 64.2 & 65.8 & 66.2 & 55.8 & 61.0 \\
Text-Only SFT (on \dataset) & 8B & SFT & 32 & 67.6 & 66.8 & 68.2 & 62.0 & 65.0 \\
\rowcolor{Budapest}
\textbf{\method{}} & 8B & SFT & 32 & 70.5 & 73.1 & 77.6 & 72.2 & \textbf{73.2} \\
\rowcolor{Budapest}
\textbf{\method{}} & 8B & SFT+RL & 32 & \textbf{83.2} & \textbf{86.5} & \textbf{86.6} & \textbf{85.1} & \textbf{85.4} \\
\bottomrule
\multicolumn{9}{l}{\footnotesize $^{\dagger}$LVR often collapses under dense video visual-token inputs and fails to produce valid text responses.} \\
\end{tabular}
}
\end{table*}

\begin{table}[t]
\centering
\footnotesize
\setlength{\tabcolsep}{3pt}
\renewcommand{\arraystretch}{1.05}
\caption{\textbf{Main results on TwiFF-Bench}~\cite{liu2026twiff}. Avg.=(CoT+Ans)/2; best results are in \textbf{bold}.}
\label{tab:twiffbench}
\resizebox{\columnwidth}{!}{
\begin{tabular}{@{}l|c|ccc@{}}
\toprule
\textbf{Model} & \textbf{Size} & \textbf{CoT} & \textbf{Answer} & \textbf{Avg.} \\
\midrule
\multicolumn{5}{@{}l@{}}{\textit{Multimodal Large Language Models}} \\
Qwen2.5-VL~\cite{bai2025qwen25vltechnicalreport} & 7B & 2.46 & 1.63 & 2.05 \\
InternVL3.5~\cite{wang2025internvl3} & 8B & 2.35 & 1.85 & 2.10 \\
DeepEyes~\cite{zheng2025deepeyes} & 7B & 2.54 & 2.20 & 2.37 \\
\midrule
\multicolumn{5}{@{}l@{}}{\textit{Unified Models}} \\
Janus-Pro~\cite{chen2025janus} & 7B & 2.04 & 1.04 & 1.54 \\
Bagel~\cite{deng2025emergingpropertiesunifiedmultimodal} & 7B & 2.29 & 1.85 & 2.07 \\
TwiFF-300K~\cite{liu2026twiff} & 7B & 2.90 & 2.55 & 2.73 \\
TwiFF-2.7M~\cite{liu2026twiff} & 7B & 2.95 & 2.62 & 2.79 \\
\midrule
\multicolumn{5}{@{}l@{}}{\textit{Ours}} \\
Zero-Shot~\cite{bai2025qwen3} & 8B & 2.75 & 2.14 & 2.44 \\
\textbf{\modelSFT} & 8B & 2.62 & 2.42 & 2.52 \\
\rowcolor{Budapest}
\textbf{\modelRL} & 8B & \textbf{3.11} & \textbf{2.97} & \textbf{3.04} \\
\bottomrule
\end{tabular}
}
% \vspace{-5mm}
\end{table}

% \subsection{Setup}

\paragraph{Benchmarks.}
We evaluate \method{} on two complementary video event prediction benchmarks. \emph{FutureBench}~\cite{wang2025fostering} is a multiple-choice VEP benchmark that asks models to predict unobserved future events from a video prefix. It reports overall accuracy and four reasoning-depth splits: \textit{1-Hop}, \textit{2-Hop}, \textit{3-Hop}, and \textit{Interp.}. While \textit{1-Hop} mainly tests immediate next-event prediction, \textit{3-Hop} and \textit{Interp.} form harder OOD-style regimes: \textit{3-Hop} requires extrapolating longer future event chains, and \textit{Interp.} requires reasoning over non-consecutive future states under partial intermediate anchors. These splits therefore test whether a model can generalize beyond local next-event cues. \emph{TwiFF-Bench}~\cite{liu2026twiff} evaluates open-ended future-frame reasoning over 1,078 QA samples and scores both the generated reasoning trajectory and the final answer. Following the official protocol, we report CoT quality, answer quality, and their average under the benchmark judge. The TwiFF-Bench evaluation set is not used in \dataset{} construction, SFT, or RL training.

\paragraph{Implementation Details.}
We use Qwen3-VL-8B-Instruct~\cite{bai2025qwen3} as the backbone. SFT trains for 1 epoch on \dataset{} (\S\ref{ssec:sft50k}) with global batch size 128, peak learning rate $1{\times}10^{-5}$, MSE weight $\lambda{=}0.1$, and maximum latent budget $L_{\max}{=}4$ unless otherwise specified. RL starts from the SFT checkpoint with group size $G{=}8$ and uses Qwen3.6-27B as the LLM-as-judge for the accuracy reward. All experiments run on $8{\times}$NVIDIA H200 GPUs, and all checkpoints are evaluated with \textit{lmms-eval}~\cite{zhang-etal-2025-lmms}. More detailed settings are listed in Appendix~\ref{app:sec:More_Implementation_Details}.

% \paragraph{Baselines.}
% We compare against four groups: \textbf{(i)} General open-source models(Qwen2.5/3-VL~\cite{bai2025qwen25vltechnicalreport,bai2025qwen3}, InternVL3, MiMo-VL, GLM-4.1V, and proprietary GPT-4o/5~\cite{openai2024gpt4o}; \textbf{(ii)} video reasoning models with text-CoT SFT or RL (NEP~\cite{wang2025fostering}, Video-CoE~\cite{su2026video}, Video-R1~\cite{feng2025videor1reinforcingvideoreasoning}, VideoChat-R1~\cite{li2025videochatr1}); \textbf{(iii)} latent visual reasoning models (LVR~\cite{li2025latent}, Monet~\cite{wang2025monet}, SwimBird~\cite{tong2026swimbird}); and \textbf{(iv)} unified vision-language generators on TwiFF-Bench (Janus-Pro, Bagel, TwiFF-Lite/2.7M~\cite{liu2026twiff}). For controlled comparison, we additionally report the zero-shot backbone Qwen3-VL-8B-Instruct and a \emph{text-only SFT} variant trained on the same top-50K source without latent supervision.

\subsection{Main Results}

\paragraph{Prior Models Struggle on VEP.}
Tables~\ref{tab:futurebench} and~\ref{tab:twiffbench} show that VEP remains difficult even for strong MLLMs. Proprietary and open-source models do not reliably solve FutureBench: GPT-4o obtains 59.0, GPT-5 obtains 57.9, and Qwen3-VL-30B-A3B reaches 66.9. Video-reasoning models improve over generic MLLMs but continue to struggle, including Video-R1 (63.3), Video-o3 (68.9), NEP (67.3), and Video-CoE (75.0). Their remaining errors are especially visible on the harder future-oriented splits: the strongest Video-CoE reaches only 71.6 on 3-Hop and 71.4 on Interp., where models must extrapolate \textbf{longer} event chains or reason over \textbf{non-consecutive} future states. Existing \textbf{static} latent visual reasoning methods also do not transfer directly to dense video prediction: Monet reaches 47.9 and LVR obtains 21.0. These results suggest that VEP is not solved by scaling generic MLLMs, adding text-centric video reasoning, or directly reusing static latent-reasoning recipes.

\paragraph{\method{} Boosts FutureBench.}
\modelSFT{} reaches \textbf{73.2}, improving the Qwen3-VL backbone (from 61.0) by \textbf{+12.2}. It outperforms the text-only SFT control trained on the same \dataset{} (65.0) by \textbf{8.2}, isolating the gain from interleaved latent reasoning rather than sample selection alone. After LA-DAPO, \modelRL{} improves to \textbf{85.4}, exceeding Qwen3-VL-30B-A3B by \textbf{18.5} points and Video-CoE by \textbf{10.4} points. The gains over the backbone are strongest on the harder splits: \textbf{+19.0}, \textbf{+20.7}, \textbf{+20.4}, and \textbf{+29.3} on 1-Hop, 2-Hop, 3-Hop, and Interp., respectively. The larger improvements on 3-Hop and Interp. suggest that latent channel generalizes to longer future chains, rather than only improving single-step NEP.

\paragraph{TwiFF-Bench Shows the Same Trend.}
On TwiFF-Bench, \modelSFT{} raises the average score from 2.44 to 2.52. Though its CoT score decreases from 2.75 to 2.62, its answer score rises from 2.14 to \textbf{2.42}, showing the curated traces strengthen prediction even when their surface reasoning is imperfect. LA-DAPO improves both dimensions, reaching \textbf{3.11} CoT and \textbf{2.97} Ans for an average of \textbf{3.04}. This surpasses the previous best TwiFF-2.7M (2.79) and all listed MLLM or unified baselines, indicating that interleaved latent reasoning and trajectory-level RL are complementary.

\subsection{Ablation Study}

\begin{table}[t]
\centering
\small
\caption{\textbf{SFT hyperparameter ablation on FutureBench.} Accuracy (\%) for latent MSE weight $\lambda$ and budget $L_{\max}$.}
\label{tab:ablation_sft}
\resizebox{.95\columnwidth}{!}{%
\begin{tabular}{c|cccc|c}
\toprule
\textbf{Setting} & \textbf{1-Hop} & \textbf{2-Hop} & \textbf{3-Hop} & \textbf{Interp.} & \textbf{AVG} \\
\midrule
\multicolumn{6}{l}{\textbf{\textit{Latent MSE weight $\lambda$}}} \\
0.01 & 68.2 & 69.9 & 73.1 & 67.5 & 69.1 \\
0.05 & 71.1 & 72.0 & 73.6 & 69.3 & 70.9 \\
\rowcolor{Budapest}
0.10 & 70.5 & 73.1 & \textbf{77.6} & \textbf{72.2} & \textbf{73.2} \\
0.20 & 69.9 & \textbf{76.7} & 74.6 & 70.1 & 72.2 \\
0.50 & \textbf{73.4} & 71.0 & 71.6 & 69.3 & 70.7 \\
1.00 & \textbf{73.4} & 73.1 & 68.7 & 67.1 & 69.5 \\
\midrule
\multicolumn{6}{l}{\textbf{\textit{Maximum latent budget $L_{\max}$}}} \\
2 & 66.5 & 74.1 & 74.6 & 69.3 & 70.7 \\
\rowcolor{Budapest}
4 & 70.5 & 73.1 & \textbf{77.6} & 72.2 & \textbf{73.2} \\
8 & 65.9 & \textbf{75.1} & 73.6 & \textbf{72.4} & 72.1 \\
16 & 69.9 & 72.5 & 71.1 & 70.8 & 71.0 \\
% \textbf{20} & 69.4 & 71.5 & 72.6 & 65.6 & 68.7 \\
% \textbf{24} & \textbf{72.8} & 67.9 & 71.6 & 65.6 & 68.4 \\
32 & 69.4 & 72.0 & 71.1 & 69.5 & 70.3 \\
64 & 67.1 & 68.9 & 70.6 & 65.6 & 67.4 \\
\bottomrule
\end{tabular}}
\end{table}

\paragraph{SFT Hyperparameters.}
Table~\ref{tab:ablation_sft} sweeps the latent MSE weight \(\lambda\) and the maximum latent budget \(L_{\max}\). With \(L_{\max}=4\) fixed, \(\lambda=0.1\) is optimal (\textbf{73.2}); both weaker (\(\lambda=0.01\), 69.1) and stronger (\(\lambda=1.0\), 69.5) alignment weights cost 3-4 points, indicating that latent positions need explicit but not dominant supervision. With \(\lambda=0.1\) fixed, accuracy peaks at \(L_{\max}=4\) and degrades to \textbf{67.4} at \(L_{\max}=64\), suggesting that an overly long latent span dilutes useful signal. This indicates that latent reasoning benefits from short, explicitly supervised spans rather than simply allocating more continuous tokens. We adopt \(\lambda=0.1\), \(L_{\max}=4\) as the default SFT setting.

\begin{table}[t]
\centering
% \small
\caption{\textbf{RL objective ablation on FutureBench.} Accuracy (\%); all variants start from \modelSFT{}.}
\label{tab:ablation_rl_method}
\resizebox{\columnwidth}{!}{%
\begin{tabular}{l|cccc|c}
\toprule
\textbf{Method} & \textbf{1-Hop} & \textbf{2-Hop} & \textbf{3-Hop} & \textbf{Interp.} & \textbf{AVG} \\
\midrule
Text-Only SFT & 67.6 & 66.8 & 68.2 & 62.0 & 65.0 \\
\quad + GRPO & 77.5 & 78.8 & 78.1 & 77.1 & 77.7 \\
\quad + DAPO & \textbf{83.2} & 81.3 & 78.1 & 71.2 & 76.3 \\
\midrule
\modelSFT{} & 70.5 & 73.1 & 77.6 & 72.2 & 73.2 \\
\quad + GRPO & 82.7 & 84.5 & 85.1 & 81.2 & 82.8 \\
\quad + DePO & 78.0 & 80.3 & 86.6 & 80.2 & 81.1 \\
\quad + DAPO & \textbf{83.2} & 85.5 & 86.6 & 82.4 & 83.8 \\
\quad \quad + \(R_{\mathrm{ctr}}\) & \textbf{83.2} & 86.0 & 87.1 & 83.2 & 84.5 \\
\quad \quad + \(R_{\mathrm{div}}\) & 82.7 & \textbf{87.0} & \textbf{87.6} & 83.4 & 84.8 \\
\rowcolor{Budapest}
\modelRL{} & \textbf{83.2} & 86.5 & 86.6 & \textbf{85.1} & \textbf{85.4} \\
\bottomrule
\end{tabular}
}
\end{table}

\begin{table}[t]
\centering
% \small
% \setlength{\tabcolsep}{3.5pt}
% \renewcommand{\arraystretch}{1.1}
\caption{\textbf{LA-DAPO reward coefficient ablation on FutureBench.} Accuracy (\%) for $\lambda_c$ and $\lambda_d$.}
\label{tab:ablation_rl_lambda}
\resizebox{.9\columnwidth}{!}{%
\begin{tabular}{c|cccc|c}
\toprule
\textbf{Setting} & \textbf{1-Hop} & \textbf{2-Hop} & \textbf{3-Hop} & \textbf{Interp.} & \textbf{AVG} \\
\midrule
\multicolumn{6}{l}{\textbf{\textit{Outcome-contrastive weight $\lambda_c$}}} \\
% 0.00 & 82.7 & \textbf{87.0} & \textbf{87.6} & 83.4 & 84.75 \\
0.01 & 81.5 & 84.5 & 86.1 & 83.4 & 83.8 \\
0.05 & 82.7 & \textbf{87.0} & 86.1 & 83.0 & 84.3 \\
0.10 & \textbf{84.4} & 86.5 & 87.1 & 84.0 & 85.1 \\
\rowcolor{Budapest}
0.20 & 83.2 & 86.5 & 86.6 & \textbf{85.1} & \textbf{85.4} \\
0.50 & 82.1 & 86.0 & 86.1 & 83.0 & 84.0 \\
1.00 & 83.8 & 86.5 & 86.6 & 84.5 & 85.1 \\
\midrule
\multicolumn{6}{l}{\textbf{\textit{Temporal diversity weight $\lambda_d$}}} \\
% 0.00 & 83.2 & 86.0 & \textbf{87.1} & 83.2 & 84.47 \\
0.01 & 83.2 & 86.5 & 86.6 & 83.8 & 84.8 \\
0.05 & \textbf{83.8} & \textbf{87.0} & 86.6 & 84.3 & 85.1 \\
\rowcolor{Budapest}
0.10 & 83.2 & 86.5 & 86.6 & \textbf{85.1} & \textbf{85.4} \\
0.20 & 80.9 & 82.9 & \textbf{87.1} & 83.2 & 83.5 \\
0.50 & 79.8 & 83.4 & 85.6 & 81.6 & 82.4 \\
1.00 & 78.0 & 82.4 & 85.6 & 81.0 & 81.6 \\
\bottomrule
\end{tabular}}
\end{table}

\paragraph{RL Objective.}
Table~\ref{tab:ablation_rl_method} ablates the RL objective from \modelSFT{}. GRPO (82.8) and DePO~\cite{10.1007/978-3-032-37592-6_35} (81.1) already lift \modelSFT{} (73.2) by about 9 points, and DAPO further reaches \textbf{83.8}. Adding latent-aware rewards improves the objective beyond DAPO: the outcome-contrastive reward \(R_{\mathrm{ctr}}\) raises performance to 84.5, the temporal-diversity reward \(R_{\mathrm{div}}\) reaches 84.8, and using both in \modelRL{} achieves \textbf{85.4}. This shows that the gain is not only from stronger RL, but from rewards that directly structure latent visual trajectories.

\paragraph{RL Reward Coefficients.}
Table~\ref{tab:ablation_rl_lambda} examines the latent-reward coefficients. The outcome-contrastive weight peaks at \(\lambda_c=0.2\) (\textbf{85.4}), and the temporal-diversity weight peaks at \(\lambda_d=0.1\); larger values hurt, dropping to \textbf{81.6} at \(\lambda_d=1.0\). This suggests that contrastive alignment and temporal diversity are both useful, but excessive pressure can push latent spans off the manifold.

\subsection{Analysis of Latent Visual Reasoning}

\begin{table}[t]
\centering
% \small
\caption{\textbf{Effect of visual-gain filtering.} FutureBench accuracy (\%) for 50K TwiFF-format SFT data.}
\label{tab:data_filter}
\resizebox{\columnwidth}{!}{%
\begin{tabular}{l|ccccc}
\toprule
\textbf{Training Set} & \textbf{1-Hop} & \textbf{2-Hop} & \textbf{3-Hop} & \textbf{Interp.} & \textbf{AVG} \\
\midrule
Zero-Shot & 64.2 & 65.8 & 66.2 & 55.8 & 61.0 \\
Random 50K & 67.6 & 68.9 & 70.1 & 67.7 & 68.4 \\
\rowcolor{Budapest}
\dataset & \textbf{70.5} & \textbf{73.1} & \textbf{77.6} & \textbf{72.2} & \textbf{73.2} \\
\bottomrule
\end{tabular}}
\end{table}

\paragraph{Visual-Gain Filtering.}
Table~\ref{tab:data_filter} controls for a key confound: whether the SFT gain comes from visual-gain selection or merely from TwiFF-style formatting. We compare our Top-50K set with a random 50K set sampled from TwiFF-2.7M under the same interleaved-format requirement and train both with the same \modelSFT{} recipe. The random set improves Qwen3-VL-8B from 61.0 to 68.4, showing that interleaved demonstrations help, but it remains 4.8 points below our visual-gain selected set (73.2). The gap persists on the harder splits, including 3-Hop (70.1 vs. 77.6) and Interp. (67.7 vs. 72.2). Thus \dataset{} improves transfer not only by exposing the model to TwiFF-style traces, but by selecting examples whose future visual hints provide measurable predictive utility.

\begin{figure}[t]
    \centering
    \includegraphics[width=\linewidth]{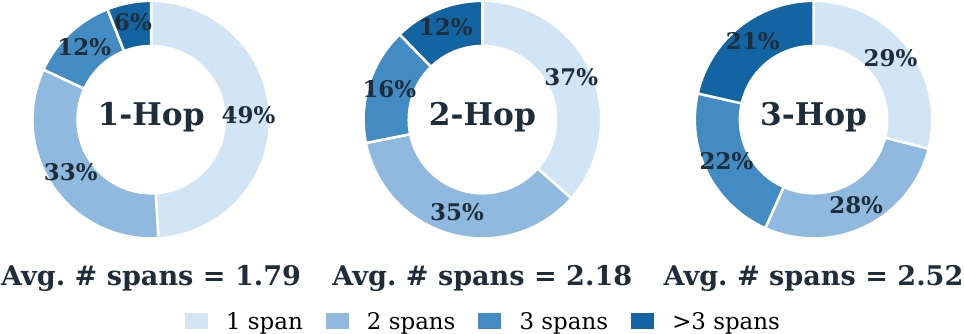}
    \caption{\textbf{Latent-span usage by reasoning depth.} Donuts show span-count distributions; values report mean spans over six RL settings.}
    \label{fig:latent_span_depth}
    \vspace{0.4em}
    \includegraphics[width=\linewidth]{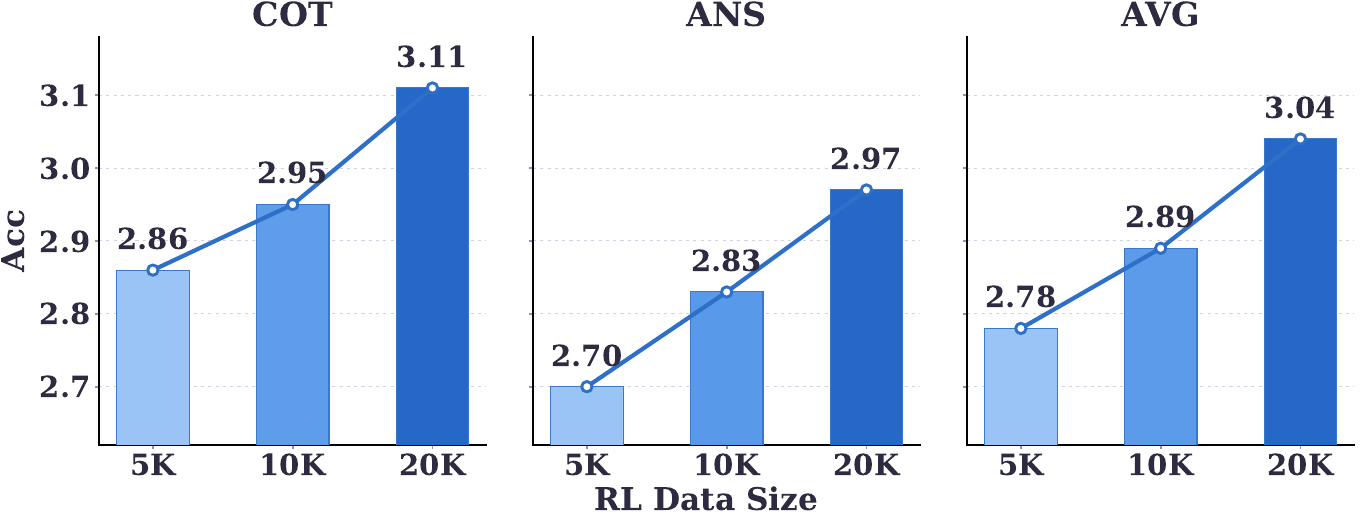}
    \caption{\textbf{RL data scaling on TwiFF-Bench.} Scores improve as LA-DAPO uses 5K, 10K, and 20K retained visual-gain samples.}
    \label{fig:data_scaling}
    \vspace{-5mm}
\end{figure}

\paragraph{Adaptive Latent Usage.}
Figure~\ref{fig:latent_span_depth} examines whether \method{} allocates latent computation according to reasoning difficulty~\cite{zhu2025llm}. Averaged over six RL hyperparameter settings, the mean span count increases with depth, from 1.79 on 1-Hop to 2.18 on 2-Hop and 2.52 on 3-Hop. The distribution shifts in the same direction: one-span responses become less frequent as depth increases, while responses with more than three spans grow from 6\% on 1-Hop to 12\% on 2-Hop and 21\% on 3-Hop. This shows that latent spans are not emitted as a fixed template; instead, \method{} spends more latent visual budgets when longer future event chains require updating dynamic visual states.

\paragraph{RL Data Scaling.}
Figure~\ref{fig:data_scaling} tests whether LA-DAPO benefits from more retained visual-gain data. Using 5K, 10K, and 20K samples from the retained pool, the TwiFF-Bench average score increases monotonically from 2.78 to 2.89 and 3.04. This trend indicates that trajectory-level latent RL continues to benefit from high-utility samples rather than saturating on a small preference set.

\begin{table}[t]
\centering
% \small
% \setlength{\tabcolsep}{6pt}
% \renewcommand{\arraystretch}{1.1}
\caption{\textbf{Inference cost on FutureBench.} Average tokens, accuracy, latency, and accuracy per second.}
\label{tab:inference_cost}
\resizebox{\columnwidth}{!}{%
\begin{tabular}{lcccc}
\toprule
\textbf{Model} & \textbf{Tokens} $\downarrow$ & \textbf{Acc.} $\uparrow$ & \textbf{Latency (s)} $\downarrow$ & \textbf{Acc./s} $\uparrow$ \\
\midrule
Video-R1 & 398.5 & 63.3 & 3.28 & 19.3 \\
Video-o3 & 348.6 & 68.9 & 25.90 & 2.7 \\
Qwen3-VL-8B & 288.8 & 61.0 & 1.18 & 51.7 \\
\textbf{\modelSFT} & 205.3 & 73.2 & 0.96 & 76.1 \\
\rowcolor{Budapest}
\textbf{\modelRL} & \textbf{195.3} & \textbf{85.4} & \textbf{0.91} & \textbf{93.8} \\
\bottomrule
\end{tabular}
}
\end{table}

\paragraph{Inference Efficiency.}
Table~\ref{tab:inference_cost} compares inference cost on FutureBench. Text-heavy and multi-turn baselines require substantially larger decoding budgets: Video-R1 emits 398.5 tokens at 3.28 seconds per sample, and Video-o3 emits 348.6 tokens at 25.90 seconds due to repeated model calls during search. In contrast, \modelSFT{} uses 205.3 tokens and reaches 73.1 accuracy at 0.96 seconds, while \modelRL{} uses 195.3 tokens and reaches 85.4 accuracy at 0.91 seconds, yielding the best accuracy-per-second score. Thus \method{} improves accuracy through compact latent visual computation rather than expensive explicit multi-turn reasoning.

More analysis including latent visualizations and reward dynamics are provided in Appendix~\ref{app:additional_analyses}.
\section{Conclusion}
\label{sec:Conclusions}

We presented \textbf{\method}, an interleaved latent visual reasoning framework for video event prediction. The central idea is to keep dynamic future visual structure in a continuous latent channel instead of verbalizing every intermediate hypothesis as text. To make this practical, \method{} first uses \textbf{\dataset} to ground latent spans with future-frame embeddings selected by visual-gain curation, and then applies \textbf{LA-DAPO} to optimize sampled latent trajectories through outcome-contrastive and temporal-diversity rewards. Across FutureBench and TwiFF-Bench, this combination improves both multiple-choice future prediction and open-ended future reasoning, with especially large gains on longer and non-consecutive future-event splits. These results suggest a broader direction for video reasoning: language should organize and communicate predictions, while latent visual states preserve the dynamic semantics needed to imagine what happens next.

% \clearpage
% Limitations section is required by the *ACL conference (placed after Conclusions, before References).
\section*{Limitations}
\method{} is developed for future-oriented video reasoning, where intermediate dynamic visual states are especially useful. Its SFT stage uses TwiFF-style trajectories with future reasoning frames; extending the same recipe to domains without such annotations may require automatic hint selection or teacher distillation. Our experiments focus on Qwen3-VL-8B and two VEP benchmarks, and future work can explore larger backbones, broader video tasks, and alternative sources of latent visual supervision. Furthermore, our evaluation data consist of instructional, predictive and English-language videos shot with static cameras, and generalization to multilingual, cinematic, or dynamically captured content remains open.

\section*{Societal Impact}
\method{} is designed for academic research on video event prediction. However, the ability to infer future behavior from video prefixes could be repurposed for surveillance systems, predictive policing, or behavior anticipation in public spaces. Erroneous predictions could lead to harmful outcomes, including false accusations or discriminatory profiling. Our current evaluation uses only publicly available research benchmarks with no real-world deployment. We encourage the community to consider responsible use guidelines when building upon future-oriented video reasoning methods.

\section*{Acknowledgments}

We thank the Shanghai Artificial Intelligence Laboratory for their institutional support. This work was supported by the Shanghai Artificial Intelligence Laboratory under Project No. PJ-PRJ24DMX001.

% Bibliography (anthology + project bib)
% {
% \small
% % \bibliographystyle{latex/acl_natbib}
% % \bibliography{anthology,reference}
% \bibliography{reference}
% }

\bibliography{custom}

\clearpage
\appendix
% \title{\MODEL: Imagine Future Events in Interleaved Latent Visual Space \\ Supplementary Materials}
% \maketitle

\section{Baselines}
\label{app:baselines}

\paragraph{General MLLMs.}
We compare against broadly trained open-source and proprietary multimodal models, including GLM-4.1V~\cite{vteam2026glm45vglm41vthinkingversatilemultimodal}, LLaVA-NeXT-Video~\cite{zhang2025llavavideovideoinstructiontuning}, MiMo-VL~\cite{coreteam2025mimovltechnicalreport}, InternVL3~\cite{zhu2025internvl3}, Qwen2.5/3-VL~\cite{bai2025qwen25vltechnicalreport,bai2025qwen3}, GPT-4o~\cite{openai2024gpt4o}, and GPT-5~\cite{singh2026openaigpt5card}. These models test whether generic video-language instruction following is sufficient for future-event prediction.

\paragraph{Video-Reasoning Models.}
We also include methods that explicitly train or optimize video reasoning behavior, including Video-RFT~\cite{wang2025videorftincentivizingvideoreasoning}, Video-R1~\cite{feng2025videor1reinforcingvideoreasoning}, VideoAuto-R1~\cite{liu2026videoauto}, Video-o3~\cite{zeng2026video}, NEP~\cite{wang2025fostering}, and Video-CoE~\cite{su2026video}. Most of these baselines use SFT, RL, or both to strengthen textual reasoning over video; they are the closest text-centric competitors to our latent visual reasoning pipeline.

\paragraph{Latent Visual Reasoning Models.}
We also compare against LVR~\cite{li2025latent}, Monet~\cite{wang2025monet}, and SwimBird~\cite{tong2026swimbird}. These models introduce non-textual or latent visual reasoning mechanisms, but were primarily developed outside dense future-event prediction. Their transfer performance helps separate the benefit of latent reasoning in general from the specific data curation and latent-aware RL used by \method{}.

\paragraph{Unified Models.}
For TwiFF-Bench, we follow the benchmark protocol and compare against representative MLLMs (Qwen2.5-VL, InternVL3.5, and DeepEyes) as well as unified understanding-generation models. Janus-Pro~\cite{chen2025janus} and Bagel~\cite{deng2025emergingpropertiesunifiedmultimodal} are unified multimodal models that support both visual understanding and generation, making them relevant baselines for future-frame reasoning beyond pure text QA. TwiFF-300K and TwiFF-2.7M~\cite{liu2026twiff} are trained on large-scale interleaved future-frame reasoning data and therefore represent the strongest TwiFF-specific unified baselines. These comparisons evaluate both the quality of the generated reasoning trajectory and the correctness of the final open-ended answer.

\begin{algorithm}[t]
\small
\caption{LA-DAPO: one update step}
\label{alg:ladapo}
\begin{algorithmic}[1]
\Require policy $\pi_\theta$, prompt $x$, group size $G$, coefficients $\lambda_{a,f,c,d}$, temperature $\tau$
\State Sample $G$ rollouts $\{y_i\}_{i=1}^G \sim \pi_{\theta_{\mathrm{old}}}(\cdot\mid x)$; record latent trajectories $\{\mathbf{Z}_i\}$
\For{$i=1,\ldots,G$}
  \State $R_{\mathrm{acc}}(i),R_{\mathrm{fmt}}(i)\gets$ rule+judge / template match on $y_i$
  \State $R_{\mathrm{ctr}}(i)\gets$ hardest-positive InfoNCE on $\mathbf{Z}_i$ vs.\ group (Eq.~\ref{eq:ctr})
  \State $R_{\mathrm{div}}(i)\gets -\tfrac{1}{M{-}1}\sum_m \cos^2(\mathbf{b}_m,\mathbf{b}_{m+1})$ (Eq.~\ref{eq:div})
  \State $R_i \gets \lambda_a R_{\mathrm{acc}}(i)+\lambda_f R_{\mathrm{fmt}}(i)+\lambda_c R_{\mathrm{ctr}}(i)+\lambda_d R_{\mathrm{div}}(i)$
\EndFor
\State $\hat{A}_i \gets (R_i - \mathrm{mean}(R))/\mathrm{std}(R)$; broadcast to text/answer tokens
\State Update $\theta$ by maximizing $\mathcal{J}_{\mathrm{DAPO}}$ with $\hat{A}_i$
\end{algorithmic}
\end{algorithm}

\begin{table*}[t]
\centering
\small
\setlength{\tabcolsep}{8pt}
\renewcommand{\arraystretch}{1.12}
\caption{\textbf{SFT hyperparameters.} Settings used to train \modelSFT{}.}
\label{tab:sft_hparams_appendix}
\begin{tabular}{@{}cc@{}}
\toprule
\textbf{Item} & \textbf{Value} \\
\midrule
Initialization & Qwen3-VL-8B-Instruct~\cite{bai2025qwen3} \\
Training data & \dataset{} \\
LLM Backbone & Full tuning \\
Vision tower / merger & Frozen \\
Precision & bf16 \\
engine & DeepSpeed ZeRO-2 \\
Optimizer & AdamW \\
$\beta_1,\beta_2$ & $0.9,0.95$ \\
Weight decay & $0.1$ \\
Gradient clip & $1.0$ \\
Schedule / warm-up & Cosine / $0.1$ \\
Peak LR & $1{\times}10^{-5}$ \\
Global batch & $128$ \\
Sequence length & $16{,}384$ \\
Frames & $16$ \\
MSE weight & $\lambda{=}0.1$ \\
Latent budget & $L_{\max}{=}4$ \\
\bottomrule
\end{tabular}
\end{table*}

\begin{table*}[t]
\centering
\small
\setlength{\tabcolsep}{8pt}
\renewcommand{\arraystretch}{1.12}
\caption{\textbf{RL / LA-DAPO hyperparameters.} Settings used to train \modelRL{}.}
\label{tab:rl_hparams_appendix}
\begin{tabular}{@{}cc@{}}
\toprule
\textbf{Item} & \textbf{Value} \\
\midrule
Initialization & \modelSFT{} checkpoint \\
Training data & FutureBench: 2K; TwiFF-Bench: 20K \\
RL framework & \textit{Easy-R1}~\cite{zheng2025easyr1} \\
% Update & DAPO token-level policy update \\
% Value function & $0$ \\
Rollout batch & $64$ \\
Group size & $G{=}8$ \\
Max prompt length & $8{,}192$ \\
Max response length & $2{,}048$ \\
Temperature / top-$p$ & $0.9 / 0.99$ \\
$\lambda_a$ & $0.9$ \\
$\lambda_f$ & $0.1$ \\
% $\lambda_c$ & $\{0.1,0.5\}$ \\
% $\lambda_d$ & $\{0.01,0.1\}$ \\
Clip & $\epsilon_l{=}0.2$, $\epsilon_h{=}0.28$ \\
Dual clip & $3.0$ \\
KL coeff. & $10^{-2}$ \\
Group filter & mean acc. $\in[0.1,0.9]$ \\
Judge model & Qwen3.6-27B \\
\bottomrule
\end{tabular}
\end{table*}

\section{LA-DAPO Details}
\label{app:ladapo}

\paragraph{DAPO Objective.}
DAPO~\cite{yu2025dapoopensourcellmreinforcement} optimizes a token-level surrogate with an asymmetric PPO clip. Let $r_t(\theta) = \pi_\theta(y_t \mid y_{<t},x)/\pi_{\theta_{\mathrm{old}}}(y_t \mid y_{<t},x)$ be the importance ratio and $\hat{A}_t$ the group-relative advantage. The per-token loss is
\begin{equation*}
\mathcal{J}_{\mathrm{DAPO}}=\mathbb{E}\big[\min(r_t \hat{A}_t,\ \mathrm{clip}(r_t,1-\epsilon_l,1+\epsilon_h)\hat{A}_t)\big],
\end{equation*}
with $\epsilon_h > \epsilon_l$ so positive-advantage tokens enjoy a wider exploration window than negative ones. We use $\epsilon_l{=}0.2$, $\epsilon_h{=}0.28$, $G{=}8$, and apply this loss only to textual and answer tokens.

\paragraph{Accuracy Reward $R_{\mathrm{acc}}\in\{0,1\}$.}
We extract the \texttt{<answer>...</answer>} span and judge it against the ground truth. Rule-based matching is tried first; the LLM judge (Qwen3.6-27B-Instruct) is invoked only on rule-failed samples, using the prompt in Figure~\ref{fig:accuracy_judge_prompt}. $R_{\mathrm{acc}}{=}1$ if either step accepts the prediction, else $0$.

\paragraph{Format Reward $R_{\mathrm{fmt}}\in\{0,1\}$.}
Following the template in Figure~\ref{fig:training_sample}, a valid response opens with a \texttt{<reason>...</reason>} block, alternates one or more \texttt{<|latent\_start|>...<|latent\_end|>} latent spans each followed by a \texttt{<reason>...</reason>}, and terminates with a single \texttt{<answer>...</answer>}. Responses with zero latent spans (pure text CoT) or any malformed boundary receive $R_{\mathrm{fmt}}{=}0$.

\paragraph{Edge Cases for $R_{\mathrm{ctr}}$.}
Group accuracy is frequently skewed early in RL, so we define $R_{\mathrm{ctr}}$ explicitly when either partition is empty: when $\mathcal{P}_i{=}\emptyset$ we set $R_{\mathrm{ctr}}(i)=1/(1+\sum_{j\in\mathcal{N}_i}\exp(s_{ij}/\tau))$, rewarding the absence of close negatives; when $\mathcal{N}_i{=}\emptyset$ we use $s_i^{+}/\tau$; when both are empty we set $R_{\mathrm{ctr}}(i)=0$.

\paragraph{Edge Cases for $R_{\mathrm{div}}$.}
When a rollout produces $M\le 1$ latent spans there are no adjacent pairs and we set $R_{\mathrm{div}}{=}0$. Span representatives with zero $L_2$ norm are skipped to avoid division-by-zero in the cosine.

\section{Implementation Details}
\label{app:sec:More_Implementation_Details}

The training hyperparameters for the SFT and LA-DAPO stages are summarized in Tables~\ref{tab:sft_hparams_appendix} and~\ref{tab:rl_hparams_appendix}, respectively. We implement the RL stage with the open-source \textit{Easy-R1} framework~\cite{zheng2025easyr1}.

\section{Additional Evaluation Details}
\label{app:evaluation_details}

\subsection{Benchmark Details}

\paragraph{FutureBench.}
FutureBench~\cite{wang2025fostering} evaluates multiple-choice video event prediction from an observed video prefix. Each example provides a video, a question, four candidate future-event continuations, and a single correct option. The benchmark separates examples by temporal reasoning depth: 1-Hop asks for the next immediate future event, 2-Hop and 3-Hop require progressively longer event chains, and Interp. requires reasoning over non-consecutive future events under partial intermediate anchors. We report overall accuracy and the four split accuracies. For RL, we follow NEP~\cite{wang2025fostering} and Video-CoE~\cite{su2026video} and train LA-DAPO for one epoch on a 2K training set.

\paragraph{TwiFF-Bench.}
TwiFF-Bench~\cite{liu2026twiff} evaluates open-ended future-frame reasoning. Each example contains input frames sampled from the observed prefix, a forecasting question, reference future reasoning with intermediate reasoning images, and a ground-truth answer. The task covers instructional, predictive, and camera-centric scenarios. Unlike FutureBench, TwiFF-Bench is not a multiple-choice benchmark: it evaluates both the model's reasoning trajectory and final answer on a 0--5 scale, and the reported score is the average of the two dimensions. For RL, we randomly sample 20K format-valid examples from the retained visual-gain pool and train for one epoch. All SFT and RL training sets are filtered to be disjoint from the reported benchmark evaluation sets, ensuring no overlap between training examples and measured test samples.

\subsection{\textit{lmms-eval} Evaluation Configuration}

For FutureBench, we evaluate each sample with up to 32 input frames and allow at most $2{,}048$ new tokens. For TwiFF-Bench, we allow at most $4{,}096$ new tokens. Both benchmarks use deterministic decoding: temperature $0$, top-$p$ $1$, beam size $1$, and sampling disabled.

\section{Details of \dataset}
\label{app:dataset_details}

\dataset{} is the 50K subset used to cold-start latent visual reasoning before LA-DAPO. It is selected from TwiFF-format interleaved trajectories by the visual-gain probe described in \S\ref{ssec:sft50k}. Each example contains a video prefix frame, one or more future reasoning frames, and an interleaved textual reasoning trace. The retained examples emphasize cases where future visual hints substantially improve prediction reliability, so the dataset targets samples for which visual imagination is empirically useful rather than merely available.

\begin{figure}[t]
\centering
\includegraphics[width=\columnwidth]{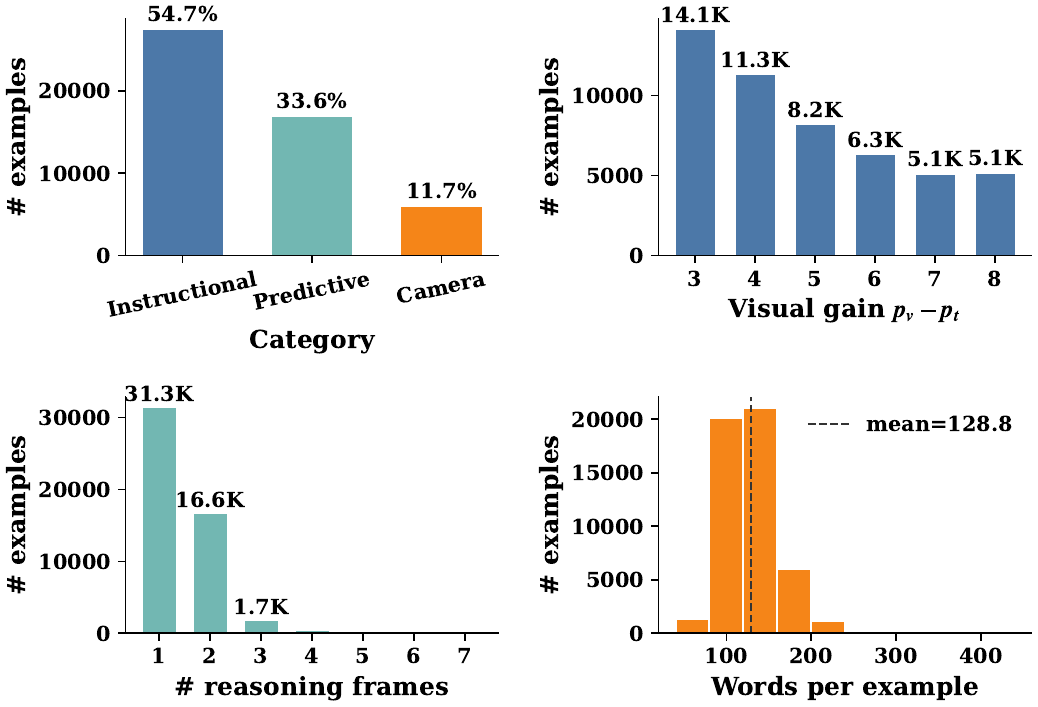}
\caption{\textbf{Statistics of \dataset{}.} Category, visual-gain, reasoning-frame count, and word-count distributions.}
\label{fig:dataset_stats_appendix}
\end{figure}

\begin{figure}[!t]
\centering
\includegraphics[width=\columnwidth]{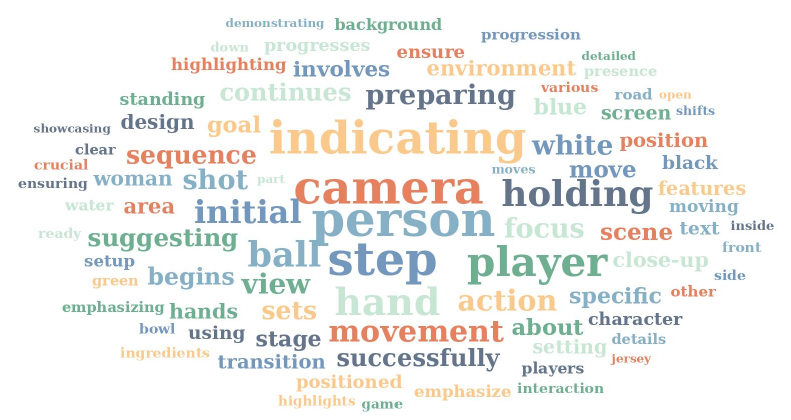}
\caption{\textbf{Word frequency in \dataset{}.}}
\label{fig:dataset_wordcloud_appendix}
\vspace{0.6em}
\includegraphics[width=.86\linewidth]{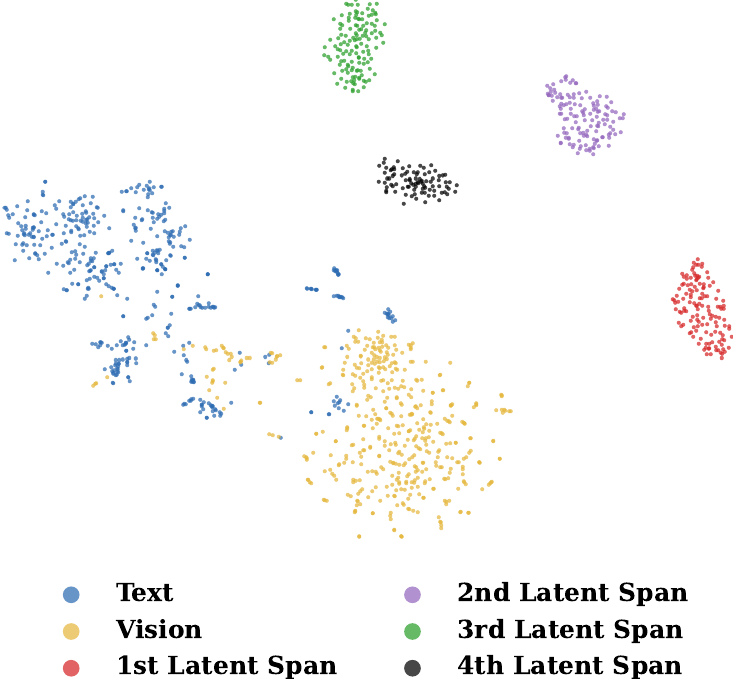}
\caption{\textbf{Stage-wise latent representation.} t-SNE of \modelRL{} embeddings on FutureBench; sequential latent spans form distinct clusters.}
\label{fig:latent_block_tsne}
% \vspace{-3mm}
\end{figure}

Figure~\ref{fig:dataset_stats_appendix} shows that \dataset{} covers all three TwiFF task categories, is dominated by high visual-gain samples. Figure~\ref{fig:dataset_wordcloud_appendix} summarizes frequent content words in the selected traces. Notably, only $4.2\%$ of \dataset{} examples contain three or more future reasoning frames, yet Figure~\ref{fig:latent_span_depth} shows that \method{} allocates three-or-more latent spans increasingly often as FutureBench depth grows. This indicates that latent usage scales with inference difficulty rather than simply mirroring the SFT trace length.

\FloatBarrier

\section{Additional Analyses}
\label{app:additional_analyses}
\begin{figure*}[t!]
    \centering
    \begin{subfigure}{0.245\linewidth}
        \centering
        \includegraphics[width=\linewidth]{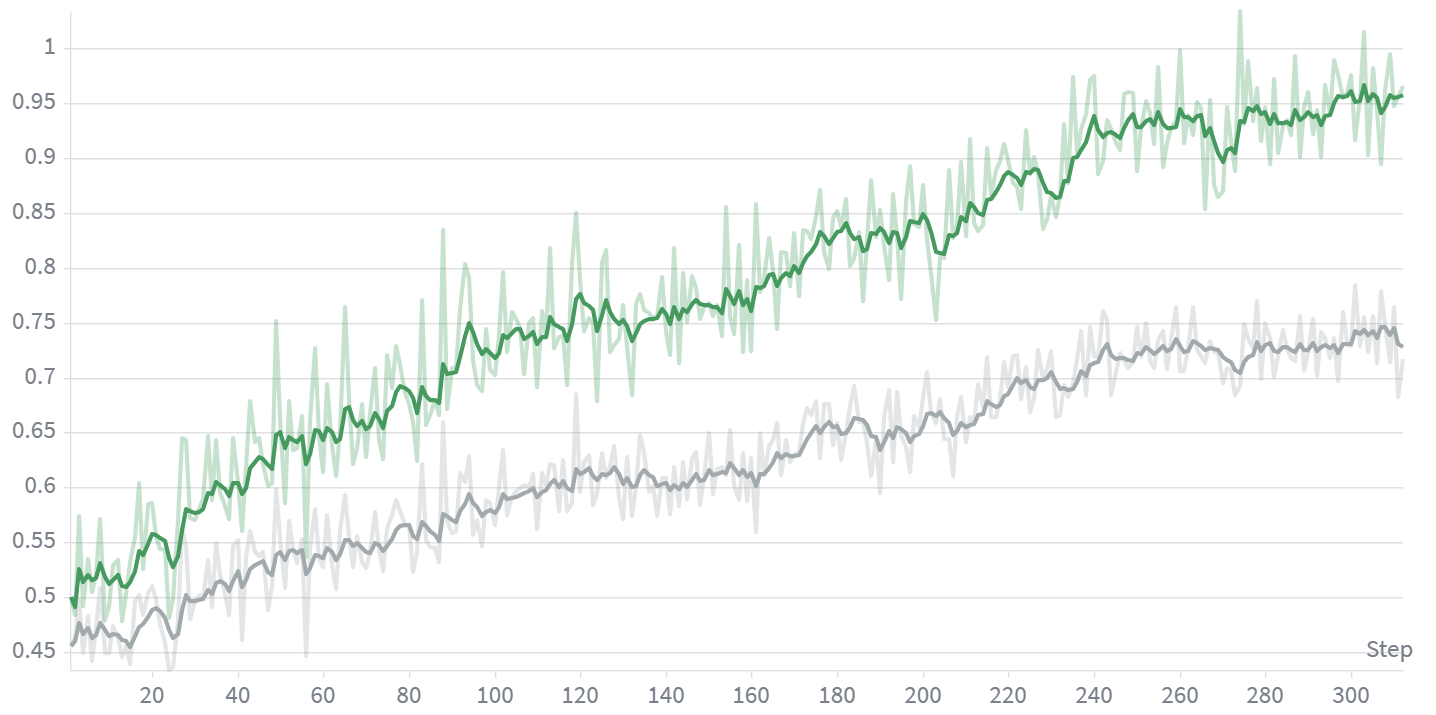}
        \caption{Overall reward}
    \end{subfigure}\hfill
    \begin{subfigure}{0.245\linewidth}
        \centering
        \includegraphics[width=\linewidth]{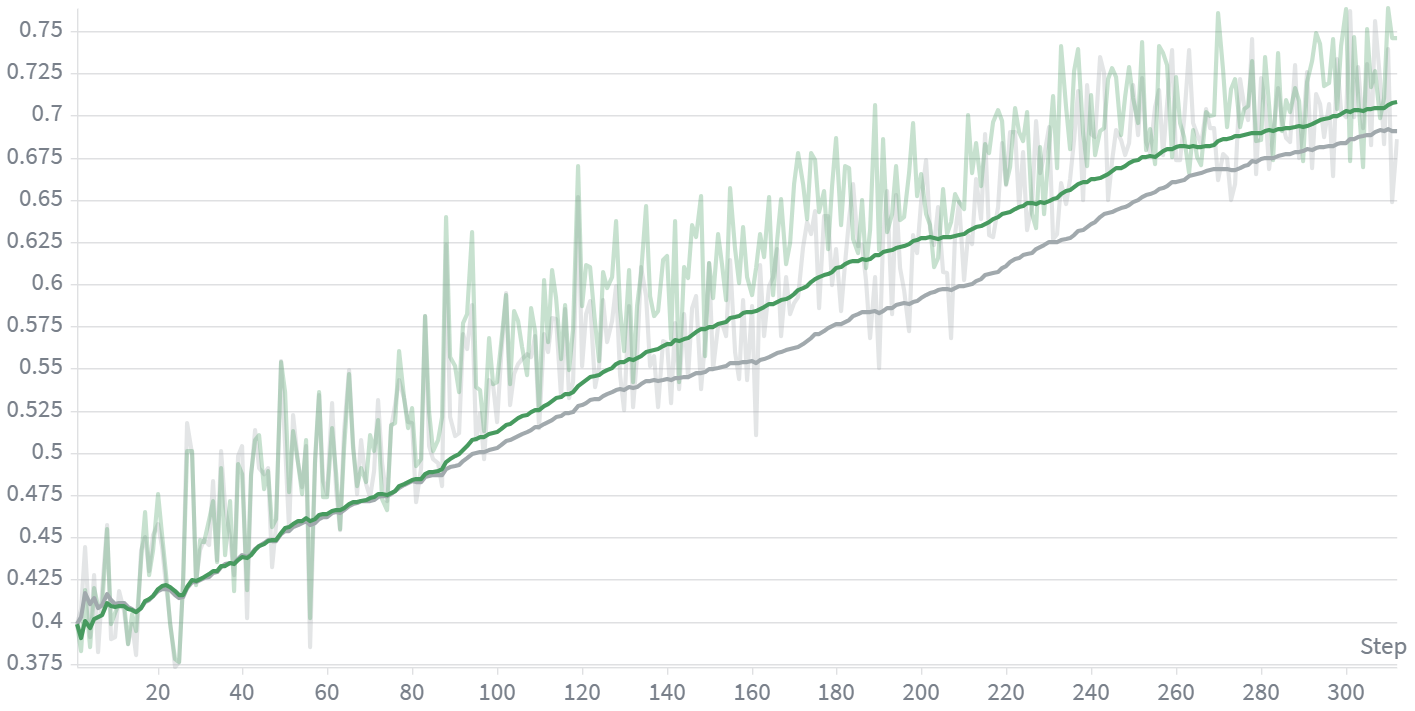}
        \caption{Accuracy reward}
    \end{subfigure}\hfill
    \begin{subfigure}{0.245\linewidth}
        \centering
        \includegraphics[width=\linewidth]{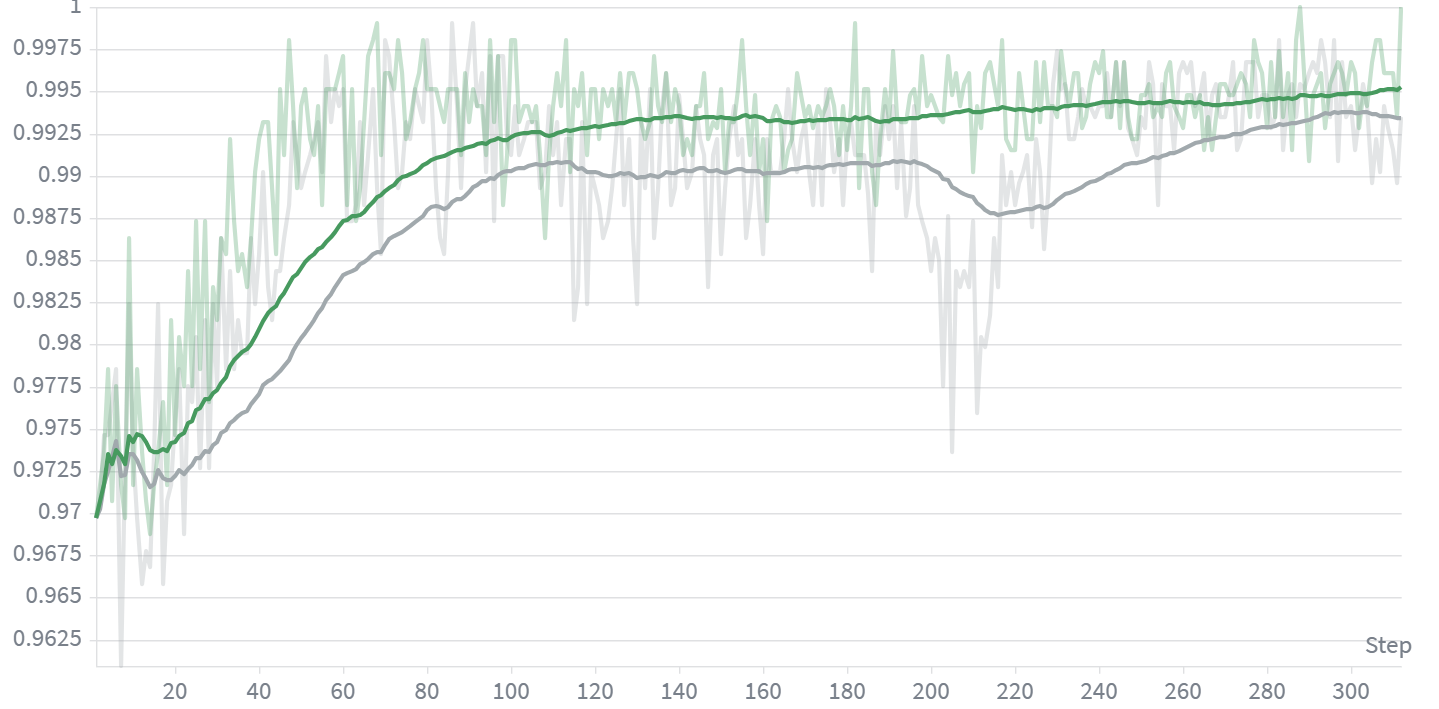}
        \caption{Format reward}
    \end{subfigure}\hfill
    \begin{subfigure}{0.245\linewidth}
        \centering
        \includegraphics[width=\linewidth]{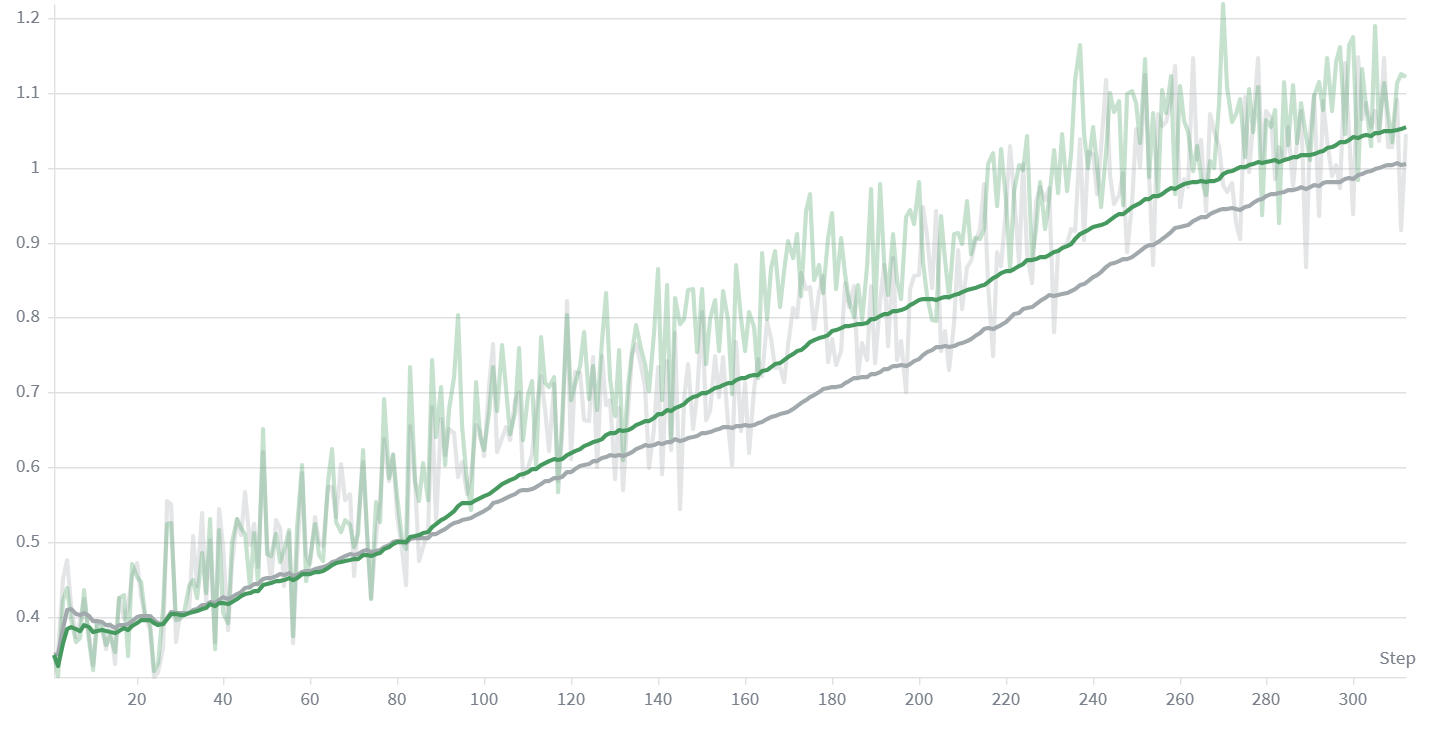}
        \caption{Contrastive visual reward}
    \end{subfigure}
    \caption{\textbf{Reward dynamics during RL.} \method{} shows higher and more stable rewards than DAPO.}
    \label{fig:reward_curve}
\end{figure*}

\paragraph{Stage-wise Latent States.}
Figure~\ref{fig:latent_block_tsne} examines whether latent spans collapse to redundant states. We visualize token embeddings from \modelRL{} on FutureBench and group latent states by span order. Text and vision tokens occupy separate modality regions, while ordered latent spans form compact clusters that are also separated from one another. This structure suggests that the model is not repeatedly emitting the same latent visual thought across time. Instead, the latent channel provides a stage-wise representation process in which successive spans update the model's internal future hypothesis before the final prediction.

\paragraph{Reward Dynamics.}
Figure~\ref{fig:reward_curve} compares the training rewards of standard DAPO and our latent-aware \method{} policy. Across the overall reward, accuracy reward, format reward, and contrastive visual reward, \method{} consistently yields higher and more stable trajectories than DAPO. The advantage is not limited to the final-answer signal: the contrastive visual reward also improves, indicating that LA-DAPO aligns latent visual states with successful prediction trajectories rather than merely optimizing textual answer format. These dynamics provide training-time evidence that the proposed latent-aware rewards make RL more effective for future-event reasoning.

\section{Prompts}
\label{app:prompts}

Figure~\ref{fig:future_l1_system_prompt} shows the system prompt that enables interleaved textual and latent visual reasoning. For TwiFF-Bench evaluation, Figure~\ref{fig:twiff_user_prompt} gives the user prompt template, while Figures~\ref{fig:twiff_judge_system_prompt} and~\ref{fig:twiff_judge_user_payload} specify the judge prompt and payload used to score reasoning quality and answer accuracy. Figure~\ref{fig:accuracy_judge_prompt} reports the binary answer-judge prompt used by the LA-DAPO accuracy reward.

\begin{figure*}[h]
\begin{tcolorbox}[
  colback=white, colframe=boxframe, colbacktitle=boxback, coltitle=black,
  arc=2pt, boxsep=2pt, left=6pt, right=6pt, top=4pt, bottom=4pt,
  title={\small \textbf{Future-L1 System Prompt}}, fonttitle=\bfseries]
\small
You are a multimodal reasoning assistant capable of thinking in textual and visual modes.\\[2pt]
Use the following tags to switch your thinking mode:\\[2pt]
1. \textbf{Textual Mode}: \texttt{<reason>Your textual reasoning process</reason>}\\
\hspace*{1em}For logical analysis, planning, and verbal thought.\\[2pt]
2. \textbf{Visual Mode}: \texttt{<|latent\_start|>Your visual reasoning process<|latent\_end|>}\\
\hspace*{1em}For mental visualization, imagination and simulation.\\[2pt]
\textbf{Output Rules}: After all thinking is complete, place the final answer inside \texttt{<answer>Your Final Answer</answer>}.
\end{tcolorbox}
\caption{\textbf{Future-L1 system prompt.}}
\label{fig:future_l1_system_prompt}
\end{figure*}

\begin{figure*}[h]
\begin{tcolorbox}[
  colback=white, colframe=boxframe, colbacktitle=boxback, coltitle=black,
  arc=2pt, boxsep=2pt, left=6pt, right=6pt, top=4pt, bottom=4pt,
  title={\small \textbf{TwiFF-Bench User Prompt Template}}, fonttitle=\bfseries]
\small
You are an AI assistant capable of reasoning with visual imagery. You should conduct a detailed analysis of the question. Consider different angles, potential solutions, and reason through the problem step-by-step with image. After fully reasoning through the problem--potentially using image-based thinking--provide only a clear, concise, and direct answer to the user's question.\\[4pt]
\{Question with \texttt{<image>} markers stripped while retaining the original frame labels\}\\[4pt]
Optional answer-tag suffix used by the prompt-suffix evaluation variant:\\
Please provide the answer within the \texttt{<answer> </answer>} tags.
\end{tcolorbox}
\caption{\textbf{TwiFF-Bench user prompt template.}}
\label{fig:twiff_user_prompt}
\end{figure*}

\begin{figure*}[h]
\begin{tcolorbox}[
  colback=white, colframe=boxframe, colbacktitle=boxback, coltitle=black,
  arc=2pt, boxsep=2pt, left=6pt, right=6pt, top=4pt, bottom=4pt,
  title={\small \textbf{TwiFF-Bench Judge System Prompt}}, fonttitle=\bfseries]
\small
You are a strict evaluator. You will have to evaluate the model response reasoning chain and answer based on the reference reasoning chain and ground truth answer.\\
Given:\\
\hspace*{1em}Question: The original forecasting question with image originates from the first video frame.\\
\hspace*{1em}Reference Reasoning Chain: What actually happened, as a reference for the rationality of the reasoning chain.\\
\hspace*{1em}Ground Truth Answer: The ground truth of the question.\\
\hspace*{1em}Model Response Reasoning Chain: The model's reasoning chain.\\
\hspace*{1em}Model Response Answer: The model's answer.\\
The rating should base on the following rules:\\
\hspace*{1em}Reasoning Chain Quality: Score 0--5 based on the logical coherence, completeness, and relevance of the reasoning (including appropriate use of multimodal information if present). The chain need not match the reference exactly but must be valid and support the final answer.\\
\hspace*{1em}Answer Accuracy: Score 0--5 based on how well the final answer matches the ground truth answer. Full credit requires correctness and completeness; partial or incorrect answers receive lower scores.\\
Put the score in a list such that output score = [score1, score2], where `score1' evaluates the Reasoning Chain and `score2' evaluates the Answer.\\
You will have to give your output in the JSON format (Keep your reasoning concise and short.):\\
\{\\
\hspace*{1em}``reasoning'': str \# the score reasoning\\
\hspace*{1em}``score'': List[int]\\
\}
\end{tcolorbox}
\caption{\textbf{TwiFF-Bench judge system prompt.}}
\label{fig:twiff_judge_system_prompt}
\end{figure*}

\begin{figure*}[h]
\begin{tcolorbox}[
  colback=white, colframe=boxframe, colbacktitle=boxback, coltitle=black,
  arc=2pt, boxsep=2pt, left=6pt, right=6pt, top=4pt, bottom=4pt,
  title={\small \textbf{TwiFF-Bench Judge User Payload Template}}, fonttitle=\bfseries]
\small
Question: \{forecasting question with the input frames inserted at the original \texttt{<image>} positions\}\\[2pt]
Reference Reasoning Chain: \{reference future-event reasoning chain, with reasoning images inserted at the original \texttt{<rimage>} positions\}\\[2pt]
Ground Truth Answer: \{ground-truth answer\}\\[2pt]
Model Response Reasoning Chain: \{sanitized model reasoning; Future-L1 latent/template tags and \texttt{<reason>} markup are removed\}\\[2pt]
Model Response Answer: \{text extracted from the model's \texttt{<answer>...</answer>} span\}
\end{tcolorbox}
\caption{\textbf{TwiFF-Bench judge user payload template.}}
\label{fig:twiff_judge_user_payload}
\end{figure*}

\begin{figure*}[h]
\begin{tcolorbox}[
  colback=white, colframe=boxframe, colbacktitle=boxback, coltitle=black,
  arc=2pt, boxsep=2pt, left=6pt, right=6pt, top=4pt, bottom=4pt,
  title={\small \textbf{Accuracy Judge System Prompt}}, fonttitle=\bfseries]
\small
You are a strict and objective answer judge. Your sole task is to determine if the model's predicted answer matches the ground-truth answer based on the question provided.\\[2pt]
\textbf{Important Rules:}\\
1. \textbf{Absolute Truth}: The ground truth is the ONLY standard. Even if you think it is factually incorrect, judge based on it.\\
2. \textbf{Multiple Choice}: Accept either the option letter (e.g., `A') or its exact content.\\
3. \textbf{Numeric/Format}: Ignore case, punctuation, and minor formatting; numeric values must be equivalent (e.g., $1.0{=}1$).\\
4. \textbf{Key Information}: For long-phrase ground truths, accept predictions capturing the essential information.\\
5. \textbf{Rubric Labels}: For short rubric/criterion ground truths, accept predictions satisfying the rubric to a human-grader standard.\\[2pt]
Output only `yes' or `no'.
\end{tcolorbox}
\caption{\textbf{Accuracy judge system prompt.}}
\label{fig:accuracy_judge_prompt}
\end{figure*}

\section{Case Study}
\label{app:case_studies}

\begin{figure*}[p]
\centering
\includegraphics[width=.92\textwidth]{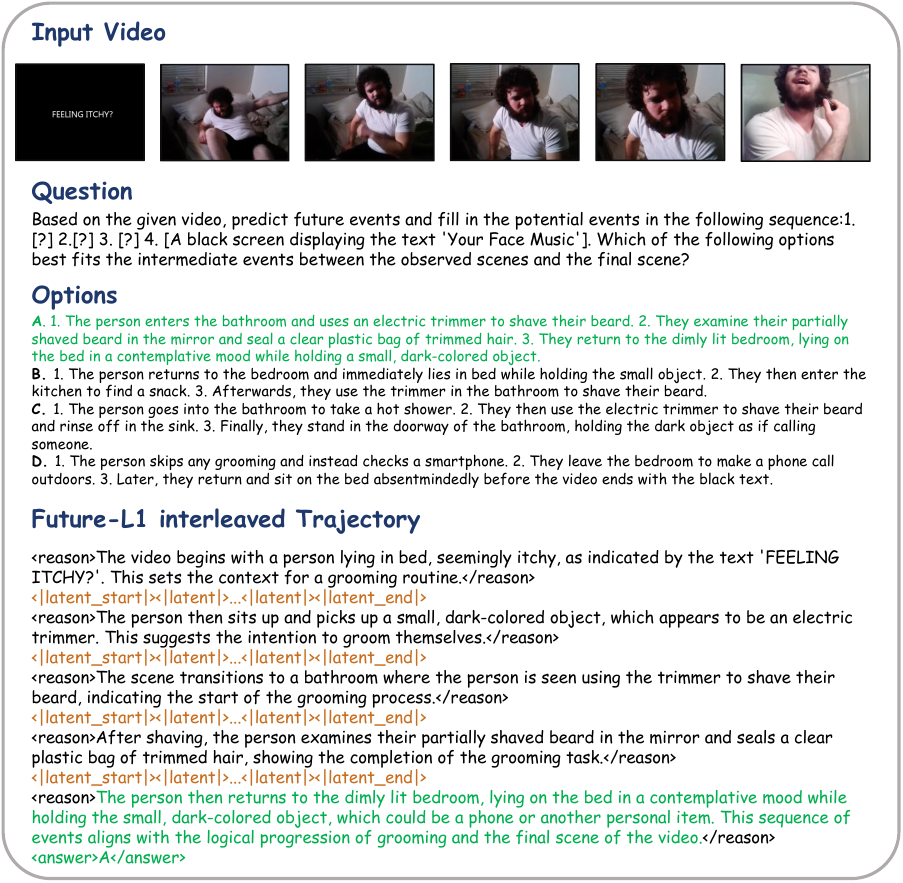}
\caption{\textbf{Successful case: grooming routine.} From an observed bedroom scene, \method{} predicts the missing sequence of beard trimming, mirror inspection, and returning to bed. The latent spans are inserted around scene and action transitions, while the text keeps the forecast interpretable.}
\label{fig:case_study_1}
\end{figure*}

\begin{figure*}[p]
\centering
\includegraphics[width=.92\textwidth]{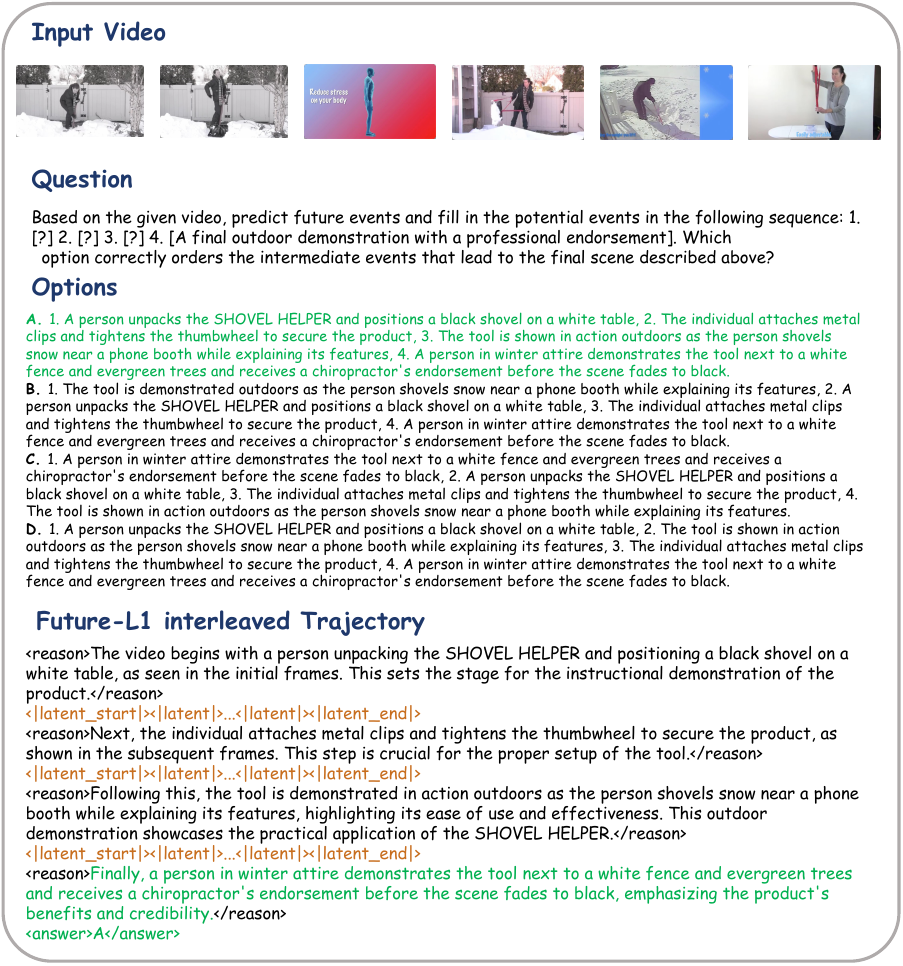}
\caption{\textbf{Successful case: product demonstration.} \method{} tracks the SHOVEL HELPER demonstration from table setup to attachment, outdoor use, and endorsement. The interleaved trajectory separates physical manipulation from later usage scenes.}
\label{fig:case_study_2}
\end{figure*}

\begin{figure*}[p]
\centering
\includegraphics[width=.92\textwidth]{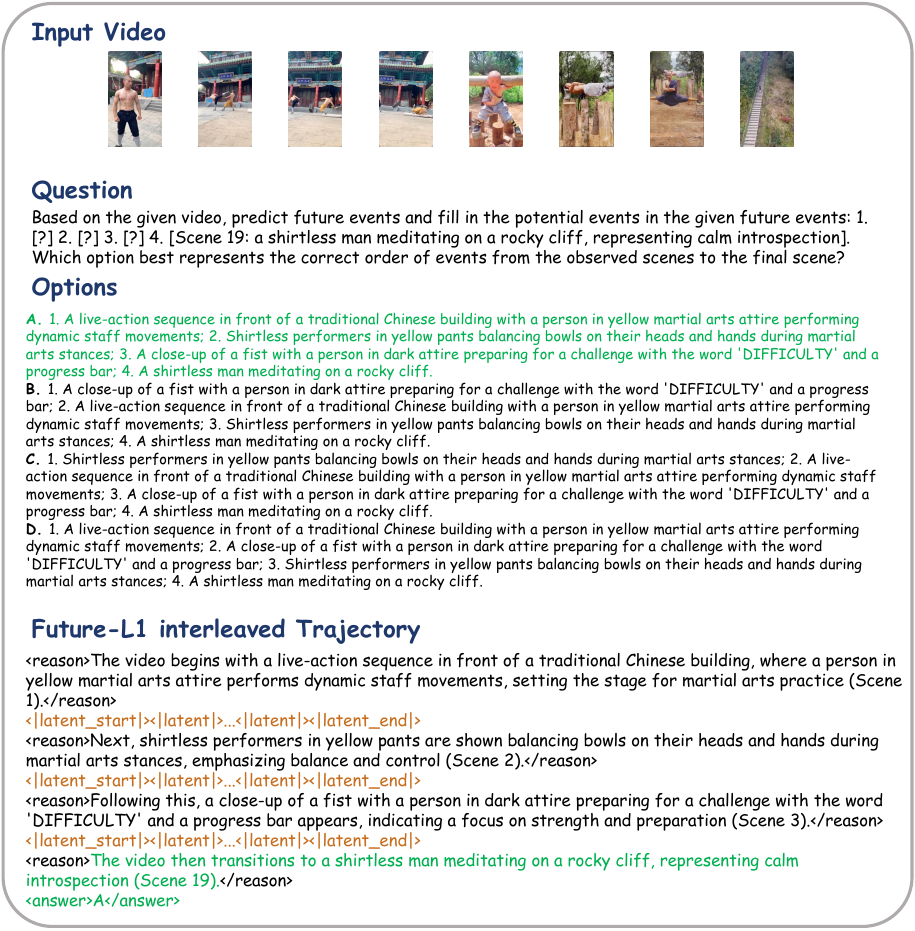}
\caption{\textbf{Successful case: staged action sequence.} \method{} follows a martial-arts montage through performance, balance practice, challenge preparation, and the final meditation scene. The latent spans help bridge visually distinct future stages before the final answer.}
\label{fig:case_study_3}
\end{figure*}

Figures~\ref{fig:case_study_1}--\ref{fig:case_study_3} provide successful qualitative examples on FutureBench. In these cases, \method{} does not compress the whole forecast into a single textual chain. Instead, it alternates short verbal anchors with latent spans at points where the future state changes: entering a new room, manipulating an object, moving from a product setup to outdoor use, or transitioning across action stages. The textual tokens make the trajectory readable, while the latent spans mark intermediate visual hypotheses that need to be carried forward before choosing the final option.

\begin{figure*}[p]
\centering
\includegraphics[width=.92\textwidth]{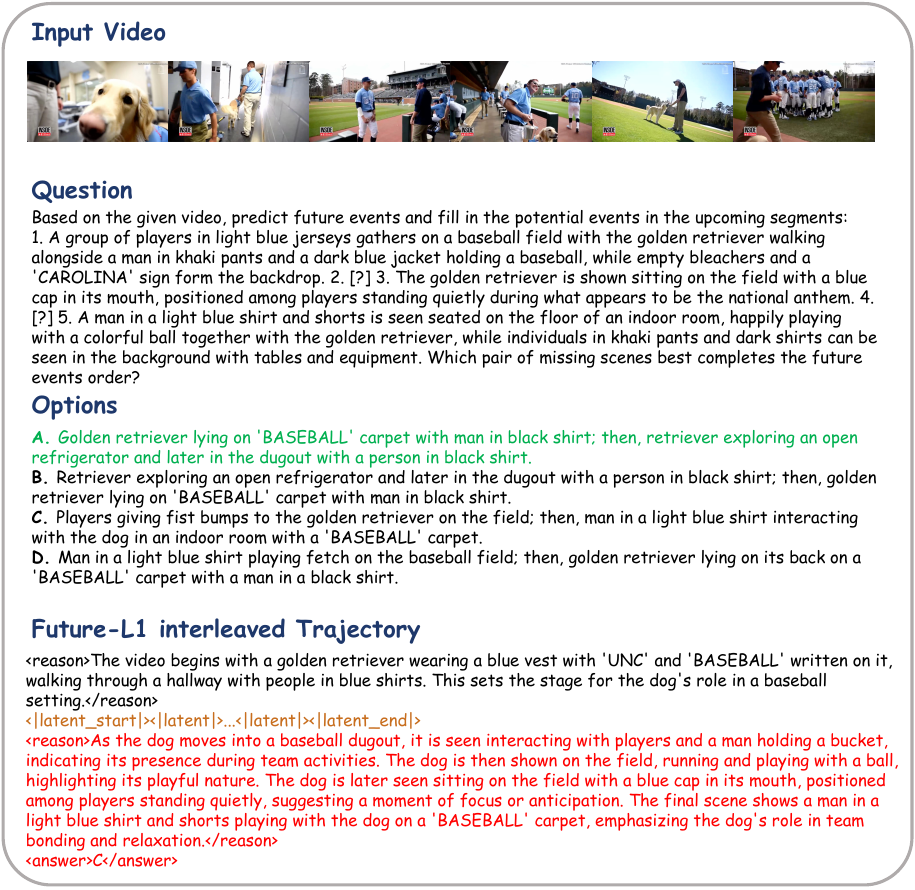}
\caption{\textbf{Failure case: event-specific detail loss.} \method{} recognizes the baseball-dog setting but predicts a generic continuation rather than the ground-truth sequence with the carpet, refrigerator, and dugout events. The example shows that latent invocation must still preserve fine-grained visual event identity.}
\label{fig:bad_case}
\end{figure*}

Figure~\ref{fig:bad_case} illustrates a representative failure. The model identifies the high-level baseball-dog context, but its latent trajectory drifts toward a plausible generic continuation and misses the specific ground-truth sequence involving the dog on the ``BASEBALL'' carpet, the open refrigerator, and the later dugout scene. This suggests that invoking latent spans is not sufficient by itself: the sampled latent trajectory must also preserve fine-grained event identity. This motivates the LA-DAPO stage, which optimizes latent trajectories with outcome-contrastive and temporal-diversity rewards.

% \clearpage
\iffalse
\section{Checklist}
\label{app:checklist}

\subsection{Artifact Licenses and Terms}
\label{app:artifact_licenses}

We use FutureBench, TwiFF-Bench, Qwen3-VL, \textit{lmms-eval}, and \textit{Easy-R1} under their respective licenses and terms. Any released code, checkpoints, or processed data will include explicit license information and will not redistribute third-party assets beyond permitted terms.

\subsection{Artifact Intended Use}
\label{app:artifact_intended_use}

We use existing artifacts only for their intended research purposes: benchmark evaluation, model initialization, training, and academic comparison. Derived artifacts are intended for research use in video event prediction and multimodal reasoning, consistent with the original access conditions.

\subsection{Artifact Documentation}
\label{app:artifact_documentation}

The artifacts cover English-language video event prediction and multimodal reasoning. FutureBench provides multiple-choice forecasting splits, TwiFF-Bench provides open-ended future-frame reasoning examples, and Appendix~\ref{app:dataset_details} documents \dataset{} statistics; we do not use or infer demographic attributes.

\subsection{Data Statistics}
\label{app:data_statistics_checklist}

We report the main data statistics in the paper and appendix: \dataset{} contains 50K examples, RL uses 2K FutureBench and 20K TwiFF-style training examples, TwiFF-Bench has 1,078 evaluation QA samples, and all training data are filtered to be disjoint from the reported evaluation sets.
\fi

\end{document}